\pdfoutput=1

\documentclass[11pt]{article}

\usepackage{EMNLP2023}

\usepackage{times}
\usepackage{array} 
\usepackage{latexsym}

\usepackage[T1]{fontenc}

\usepackage[utf8]{inputenc}

\usepackage{times}
\usepackage{latexsym}
\usepackage{xtab} 
\usepackage{tabularx}
\usepackage{graphicx}
\usepackage{amsmath,amssymb}
\usepackage{subfig}
\usepackage{stmaryrd}
\usepackage{multirow}
\usepackage{longtable}
\usepackage{tabularray}
\usepackage{tikz}
\usepackage{booktabs}
\usepackage{subcaption}
\usepackage{colortbl}
\usepackage{caption}
\usepackage{titlesec}
\usepackage{booktabs}
\usepackage{pgfplots}
\pgfplotsset{compat=1.18}
\usepackage{sansmath}
\usepackage{pifont}
\usepackage{enumitem}
\usepackage{arydshln}
\usepackage{pgfplotstable}
\pgfplotsset{compat=1.18}
\usetikzlibrary{patterns,patterns.meta}
\usepackage[dvipsnames]{xcolor}

\newif\ifcomments
\commentstrue
\ifcomments
    \providecommand{\sa}[1]{{\protect\color{red}{[SA: #1]}}} 
    \providecommand{\am}[1]{{\protect\color{magenta}{[AM: #1]}}} 
\else
    \providecommand{\sa}[1]{}    
    \providecommand{\am}[1]{}    
\fi

\commentstrue
\ifcomments
    
    \def \ifempty#1{\def\temp{#1} \ifx\temp\empty }

\usepackage{inconsolata}

\newcommand{\cmark}{\ding{51}}%
\newcommand{\xmark}{\ding{55}}%

\title{Comparing Human and Language Models Sentence Processing Difficulties on Complex Structures
}

\begin{document}

\author{Samuel Joseph Amouyal$^{\lambda}$ \hspace{0.5cm} Aya Meltzer-Asscher$^{\dagger,\star}$ \hspace{0.5cm} Jonathan Berant$^{\lambda}$ \\ 
$\lambda$ Blavatnik School of Computer Science, Tel Aviv University, Israel \\
$\dagger$ Department of Linguistics, Tel Aviv University, Israel \\
$\star$ Sagol School of Neuroscience, Tel Aviv University, Israel \\
\texttt{\{samuel.amouyal, joberant\}.cs.tau.ac.il} \\
\texttt{ameltzer@tauex.tau.ac.il}} 

\maketitle
\begin{abstract}

Large language models (LLMs) that fluently converse with humans are a reality -- but do LLMs experience human-like processing difficulties?
We systematically compare human and LLM sentence comprehension across seven challenging linguistic structures. We collect sentence comprehension data from humans and five families of state-of-the-art LLMs, varying in size and training procedure in a unified experimental framework. 
Our results show LLMs overall struggle on the target structures, but especially on garden path (GP) sentences. Indeed, while the strongest models achieve near perfect accuracy on non-GP structures (93.7\% for GPT-5), they struggle on GP structures (46.8\% for GPT-5). Additionally, when ranking structures based on average performance, rank correlation between humans and models increases with parameter count. For each target structure, we also collect data for their matched baseline without the difficult structure. Comparing performance on the target vs. baseline sentences, the performance gap observed in humans holds for LLMs, with two exceptions: for models that are too weak performance is uniformly low across both sentence types, and for models that are too strong the performance is uniformly high.
Together, these reveal convergence and divergence in human and LLM sentence comprehension, offering new insights into the similarity of humans and LLMs.\footnote{We release our code and data in \url{https://github.com/samsam3232/comparing_humans_llms_processing_difficulties}}

\end{abstract}
\section{Introduction}

The astounding advances in performance of large language models (LLMs) \cite{gpt4, geminiteam2024gemini, falcon_series, llama3herdmodels, qwen3} have sparked interest in comparing sentence processing mechanisms in humans and LLMs, with two main areas of research. The first compares brain and LLM activations while processing a sentence \cite{fedorenko_brain_corr, cacheteux-middle-layer, Goldstein2023CorrespondenceBT, Ren2024DoLL}. The second compares humans and LLMs processing outcomes, either by predicting human-related metrics (such as reading time, eye gaze, and plausibility judgments) from LLM-derived information or by directly comparing outputs on the same task \cite{linzen-etal-2016-assessing, acceptability_nn, hu-etal-2020-systematic, pretesting, Rego2024LanguageMO, Sun2024ComputationalSM, kuribayashi2025large}.

One way to estimate the similarity of humans and LLMs is to compare their error patterns. Psycholinguistic research has revealed many structures challenging humans cognitive mechanisms \cite{double_center_chomsky, frazier_newer, gibson_interference, christianson2001, christianson2006, memory-interference, zhang2023noisy}. Some prior studies have shown that LLMs struggle similarly to humans on some syntactic structures; \citet{amouyal-etal-2025-lm} found that LLMs make comprehension errors akin to humans on a specific type of ``garden path'' (GP) sentences. \citet{bert_gp} have shown that on another type of GP, LLMs struggle, but make mistakes dissimilar from humans. \citet{align_center_embedding} have shown that similar to humans, LLMs consider center-embedded sentences as ungrammatical, while \citet{sparks_double_center} has shown that when given examples, LLMs achieve almost perfect performance on these sentences.

Still, important gaps remain. First, most studies use indirect metrics (reading time for humans or surprisal for LLMs) as a proxy for processing difficulty, with few studies measuring sentence comprehension directly \cite{ferreira2019problem}. Second, LLM studies looking at specific structures used different experimental setups: different models, different data, different prompts; it is therefore hard to draw from these a unified conclusion. 

In this study, we fill these gaps by systematically comparing human and LLMs sentence comprehension, having them complete the same task: given a sentence, \emph{answer a comprehension question} on that sentence. We test the comprehension of both humans and a large family of LLMs on seven different structures, challenging different components of humans cognition.

We investigate the following structures (see Table \ref{tab:sets_examples}):
Four different types of \textbf{Garden path} (GP) sentences (e.g., \emph{The chef hired last month worked overtime.}, \emph{Did the chef hire someone?}): GP sentences are challenging because they require reanalysis of an initial parse of the sentence.
\textbf{Double center embedded} sentences (e.g., \emph{The boy that the cat the dog scared liked laughed. Who did the dog scare?}) are challenging due to multiple open dependencies items. 
\textbf{Similarity-based interference} appears when two noun phrases share some feature, which leads to impaired comprehension (e.g., \emph{The banker that the barber praised climbed the mountain. Who climbed the mountain?}).
\textbf{Depth charge} sentences (e.g., \emph{No head injury is too trivial to be ignored. Can you ignore head injuries?}) are challenging due to multiple negations.

We start by creating sentence-question pairs for all types of structures above. Each set in our data has two sentences: one with the difficult structure (the \emph{target} sentence), and a matched \emph{baseline} without the difficulty. 
For each sentence-question pair, LLMs and humans perform the same task: answer the question given the sentence. Each human participant sees only one sentence-question-pair. We test 31 state-of-the-art LLMs from 5 families, varying in size and training procedure. 

Human performance validates that these structures are challenging, where the highest average human accuracy on a task is 41.7\%. LLMs overall perform better than humans, but still struggle on these structures. The lowest mean accuracy across tasks is 28.3\%, and the highest mean accuracy is 65.3\%. Given the simplicity of the task, these results show these sentences are challenging for LLMs as well.  
In addition, reasoning tokens (`thinking') improve model performance, but only when the base model is strong enough.

\begin{table*}[t!]
\centering
\scriptsize 
\begin{tabularx}{\textwidth}{@{} l X X l l @{}}
\toprule
\textbf{Type} & \textbf{Target Sentence} & \textbf{Baseline Sentence} & \textbf{Question} & \textbf{Answer}\\
\midrule
Subject/Object (GP) 
& While the man hunted the deer ran into the woods. 
& The deer ran into the woods while the man hunted.
& Did the man hunt the deer? & No \\ 
\addlinespace 
NP/S (GP)
& The policeman saw the lights were off. 
& The policeman saw that the lights were off. 
& Did the policeman see the lights? & No \\ 
\addlinespace
NP/VP (GP)
& The complex houses married soldiers. 
& The complex housed married soldiers. 
& Are there complex houses? & No \\ 
\addlinespace
Reduced relative (GP)
& The chef hired last month worked overtime.
& The chef which was hired last month worked overtime. 
& Did the chef hire someone? & No \\ 
\midrule 
Double-center 
& The man that the teacher that the student liked called sat.
& The student liked the teacher that called the man that sat.
& Who did the student like? & The teacher \\ 
\addlinespace
Depth charge
& No head injury is too trivial to be ignored.
& Every head injury is severe enough to be ignored.
& Can you ignore head injuries? & Yes \\ 
\addlinespace
Interference
& The banker that the barber praised climbed the mountain.
& The banker that you praised climbed the mountain.
& Did the barber/you climb the mountain? & No \\
\bottomrule
\end{tabularx}
\caption{Example sentences for all constructions tested in this work.}
\label{tab:sets_examples}
\end{table*}

A key contribution of our work is to analyze the similarity between human and LLM performance, with three main findings.
\begin{enumerate}[leftmargin=*,nosep]
\item \textbf{LLMs are more similar to humans on GP structures} Comparing the absolute performance of humans and LLMs, the average absolute difference is lower for GP structures (0.17) compared to other structures (0.35). This is because LLM absolute accuracy on GP structures is \emph{lower}. Additionally, thinking tokens are less helpful on GP structures compared to the other constructions. Overall, there is a clear difference between LLM behavior on GP sentences and other phenomena. We conjecture this is due to the type of difficulty GP sentences pose: while the remaining structures stress working memory (which is larger in LLMs than in humans), GP sentences require discarding a wrong interpretation from memory. Proving this conjecture requires validation in future work.
\item \textbf{Rank correlation of the difficulty of structures between humans and LLMs increases with model size.} Spearman rank correlation between the relative ranking of average performance of humans and LLMs on the different structures  (see Figure \ref{fig:spearman_correlation}) shows that as models have more parameters, correlation increases. The model with the highest correlation is \emph{o4-mini} with a correlation of 0.93. 
\item \textbf{LLM performs better on baseline sentences when the LLM is not too strong and not too weak.} A key property of similarity between humans and LLMs is \emph{directionality}: we expect the accuracy on \emph{target}  sentences to be lower than their \emph{baseline} counterpart. Models reproduce this pattern, except in two cases. If a model is too weak, it performs poorly on both sentence types and violates directionality; if too strong, it performs equally well on both sentence types violating again directionality. The model capacity required to capture human-like directionality varies by structure, depending on how difficult that structure is for humans.
\end{enumerate}

To summarize, our contributions are:
\begin{enumerate}[nosep,leftmargin=*]
    \item We collect sentence comprehension human data on seven challenging structures
    \item We test LLM sentence comprehension on these structures in a wide array LLM families, sizes and training regimens.
    \item We provide an in-depth comparison of the similarity of human and LLM sentence processing with the aforementioned insights. 
\end{enumerate}
To our knowledge, this is the first study to examine sentence comprehension across such a large number of phenomena \emph{and} LLMs. This scale allowed us to draw broader generalizations compared to prior work.

\section{Structures examined}
\label{sec:data_used}

We examined \textbf{seven} distinct types of structures, chosen to span a range of processing difficulties. Each type challenges comprehension in a different way—through structural ambiguity, memory load, or logical/semantic complexity. 
For each structure we have two sentences: one with the difficult structure (the \emph{target} condition) and a baseline sentence where the difficulty is neutralized (the \emph{baseline} condition). Each participant sees one of the two. In both cases the sentence is followed by one comprehension question. 

We now describe each structure, providing examples of sentences and probing questions. Table ~\ref{tab:sets_examples} shows example sets for all structures.

\subsection{Garden-path sentences}

Garden‐path (GP) sentences are constructions that lead the reader into an initial incorrect parse, requiring reanalysis to achieve the correct interpretation \cite{gp3, gp2, gp1}. Such reanalysis was shown to be hard for humans \cite{christianson2001, christianson2006} . We include \textbf{four} GP subtypes:
\begin{itemize}[leftmargin=*,nosep]
    \item Subject/Object: the subject of the main verb is initially misparsed as the object of a verb in a preceding adverbial clause.
    \item NP/S: a noun phrase (NP) is initially misparsed as the object of a main clause and not as the subject of an embedded sentence (S).
    \item NP/VP: a verb (`houses') is initially misparsed as part of an NP.
    \item Reduced relative: a reduced relative clause verb is initially parsed as the main verb.
\end{itemize}

\subsection{Double center-embedded sentences}

Double-center embedded sentences are sentences with two nested clauses within another. Understanding such sentences is difficult due to the strain they impose on working memory, and they are often misunderstood or judged as ungrammatical \cite{double_center_chomsky, frazier_newer, gibson_interference, vasishth2010short}.
In each set for this structure, the question targets the deepest embedded clause.

\begin{table*}[t!]
\centering
\scriptsize
\renewcommand{\arraystretch}{1}
\begin{tabular}{l|c|ccccccc|cc}
\toprule
Model & Average & Subj/Obj & NP/VP & Depth charge & NP/S & Double center & Interference & Red. relative & GP & non GP \\
\hline 
Human & 28.3 & 13.3 & 18.5 & 28.0 & 29.7 & 32.3 & 36.9 & 41.7 & 25.8 & 32.4 \\ \hline 
o3 & 74.5 & 49.0 & 66.0 & 64.0 & 56.0 & 98.0 & 100.0 & 95.0 & 66.5 & 87.3 \\
GPT-4.1 & 68.7 & 40.0 & 63.0 & 63.0 & 57.0 & 90.0 & 100.0 & 76.0 & 59.0 & 84.3 \\
GPT-5 & 65.6 & 32.0 & 45.0 & 83.0 & 45.0 & 98.0 & 100.0 & 65.0 & 46.8 & 93.7 \\ \hline
Llama-11B-Ins. & 60.6 & 50.0 & 72.0 & 33.0 & 62.0 & 61.0 & 69.0 & 79.0 & 65.7 & 54.3 \\
Llama-11B & 59.6 & 47.0 & 66.0 & 38.0 & 56.0 & 62.0 & 82.0 & 72.0 & 60.3 & 60.7 \\
Llama-90B & 49.4 & 25.0 & 40.0 & 43.0 & 29.0 & 88.0 & 93.0 & 39.0 & 33.3 & 74.7 \\ \hline
DeepSeek-7B & 48.0 & 32.0 & 41.0 & 78.0 & 44.0 & 25.0 & 68.0 & 53.0 & 42.5 & 57.0 \\
DeepSeek-1.5B & 46.9 & 38.0 & 39.0 & 69.0 & 39.0 & 59.0 & 43.0 & 40.0 & 39.0 & 57.0 \\
DeepSeek-14B & 38.6 & 15.0 & 25.0 & 74.0 & 7.0 & 66.0 & 65.0 & 25.0 & 18.0 & 68.3 \\ \hline
Qwen-0.6B & 47.7 & 36.0 & 65.0 & 51.0 & 39.0 & 52.0 & 52.0 & 40.0 & 45.0 & 51.7 \\
Qwen-8B & 47.5 & 28.0 & 42.0 & 56.0 & 33.0 & 66.0 & 77.0 & 38.0 & 35.3 & 66.3 \\
Qwen-14B & 45.1 & 12.0 & 22.0 & 79.0 & 10.0 & 79.0 & 91.0 & 34.0 & 19.5 & 83.0 \\ \hline
Gemma-1B & 44.4 & 40.0 & 43.0 & 65.0 & 38.0 & 44.0 & 41.0 & 39.0 & 40.0 & 50.0 \\
Gemma-1B-Ins. & 39.7 & 40.0 & 52.0 & 64.0 & 34.0 & 20.0 & 28.0 & 37.0 & 40.7 & 37.3 \\
Gemma-4B & 38.6 & 35.0 & 38.0 & 55.0 & 37.0 & 35.0 & 35.0 & 34.0 & 36.0 & 41.7 \\ \hline
\bottomrule
\end{tabular}
\caption{Average accuracy of humans and LLMs by structure on the \emph{target} condition. The two rightmost columns represent respectively the average of the GP and nonGP conditions. For each family, the models are ranked in decreasing order of average accuracy. The structures are ranked in increasing order of accuracy for humans.}
\label{tab:model_accuracy}
\end{table*}

\subsection{Depth-charge sentences}

Depth-charge sentences involve multiple negations, such as \emph{``No head injury is too trivial to be ignored''}. People interpret this sentence as \emph{``you can't ignore head injuries''} (which is semantically more plausible), while the literal meaning is \emph{``you can ignore every head injury''} \cite{depth_charge, zhang2023noisy, paape2023role}.

\subsection{Similarity-based interference}

Similarity-based interference occurs when two nouns in a sentence share features (semantic, syntactic, or both), leading to interference between the two nouns during memory encoding or retrieval \cite{jager2015teasing, villata2018encoding,saul2025interference}, which can lead to comprehension problems. We use the sentences from \citet{memory-interference}, in which the interference is between two NPs that are both descriptive (e.g \emph{the cook}).

\subsection{Materials}

For Subject/Object GP, we use 45 sets from \citet{amouyal-etal-2025-lm}, along with their human results. For interference, we used 24 sets from \citet{memory-interference}, but we did not use their human results. For the remaining structures, we built 40 sets constructed as explained above. Our materials can be found in Appendix \ref{app:materials}.

\section{Experiments}

We measure the comprehension of sentences in both humans and LLMs. This allows  directly comparing humans and LLMs on the exact same task.

\begin{table*}[h!]
\scriptsize
\centering
\begin{tabular}{lc|cccc|ccc}
\hline
Model & Average & Subj/Obj & NP/VP & NP/S & Reduced relative & Depth charge & Double center & Interference \\
\hline
o3 & \cellcolor{green!10}+8.3 & \cellcolor{green!20}+21.0 & \cellcolor{green!10}+12.0 & \cellcolor{green!20}+16.0 & \cellcolor{green!10}+2.0 & \cellcolor{green!10}+8.0 & +0.0 & \cellcolor{red!10}-1.0 \\
o3-mini & \cellcolor{green!10}+2.4 & \cellcolor{green!20}+13.0 & \cellcolor{green!10}+3.0 & \cellcolor{green!10}+6.0 & \cellcolor{red!10}-4.0 & \cellcolor{green!10}+2.0 & \cellcolor{red!10}-3.0 & +0.0 \\
o4-mini & \cellcolor{green!10}+2.3 & \cellcolor{green!10}+2.0 & \cellcolor{green!10}+3.0 & \cellcolor{red!10}-3.0 & \cellcolor{green!10}+3.0 & \cellcolor{red!10}-2.0 & \cellcolor{green!20}+13.0 & +0.0 \\
GPT-5 & \cellcolor{green!20}+23.6 & \cellcolor{green!50}+55.0 & \cellcolor{green!30}+36.0 & \cellcolor{green!40}+47.0 & \cellcolor{green!30}+29.0 & \cellcolor{red!10}-1.0 & +0.0 & \cellcolor{red!10}-1.0 \\ \hline
Qwen-32B & \cellcolor{green!20}+17.9 & +0.0 & \cellcolor{green!10}+5.0 & \cellcolor{green!40}+44.0 & \cellcolor{green!30}+25.0 & \cellcolor{green!20}+19.0 & \cellcolor{green!20}+19.0 & \cellcolor{green!20}+13.0 \\
Qwen-14B & \cellcolor{green!10}+11.1 & \cellcolor{green!10}+2.0 & \cellcolor{green!10}+6.0 & \cellcolor{green!30}+32.0 & \cellcolor{green!30}+35.0 & \cellcolor{red!10}-2.0 & \cellcolor{red!10}-4.0 & \cellcolor{green!10}+9.0 \\
Qwen-8B & \cellcolor{green!10}+6.6 & \cellcolor{red!30}-27.0 & \cellcolor{red!20}-19.0 & \cellcolor{green!20}+17.0 & \cellcolor{green!20}+23.0 & \cellcolor{green!10}+9.0 & \cellcolor{green!20}+20.0 & \cellcolor{green!20}+23.0 \\
Qwen-4B & \cellcolor{green!20}+13.9 & \cellcolor{red!10}-3.0 & \cellcolor{red!20}-14.0 & \cellcolor{green!20}+22.0 & \cellcolor{green!20}+17.0 & \cellcolor{green!10}+10.0 & \cellcolor{green!30}+33.0 & \cellcolor{green!30}+32.0 \\
Qwen-1.7B & \cellcolor{green!10}+6.0 & \cellcolor{red!10}-9.0 & \cellcolor{red!20}-19.0 & \cellcolor{green!10}+3.0 & \cellcolor{red!20}-20.0 & \cellcolor{green!20}+14.0 & \cellcolor{green!30}+35.0 & \cellcolor{green!40}+38.0 \\
Qwen-0.6B & \cellcolor{red!30}-25.7 & \cellcolor{red!30}-35.0 & \cellcolor{red!50}-51.0 & \cellcolor{red!30}-28.0 & \cellcolor{red!30}-34.0 & \cellcolor{green!40}+41.0 & \cellcolor{red!30}-27.0 & \cellcolor{red!40}-46.0 \\ \hline
Deepseek-14B & \cellcolor{green!10}+4.7 & \cellcolor{red!10}-12.0 & \cellcolor{red!20}-16.0 & \cellcolor{green!10}+7.0 & \cellcolor{green!20}+21.0 & \cellcolor{red!20}-22.0 & \cellcolor{green!20}+22.0 & \cellcolor{green!30}+33.0 \\
Deepseek-7B & \cellcolor{red!10}-11.3 & \cellcolor{red!30}-30.0 & \cellcolor{red!30}-26.0 & \cellcolor{red!30}-29.0 & \cellcolor{red!20}-17.0 & \cellcolor{red!30}-26.0 & \cellcolor{green!50}+61.0 & \cellcolor{red!10}-12.0 \\
Deepseek-1.5B & \cellcolor{red!20}-17.6 & \cellcolor{red!20}-24.0 & \cellcolor{red!20}-24.0 & \cellcolor{red!30}-33.0 & \cellcolor{red!30}-28.0 & \cellcolor{green!10}+1.0 & \cellcolor{green!10}+4.0 & \cellcolor{red!20}-19.0 \\
\hline
\end{tabular}
\caption{Thinking models: Relative change resulting from enabling thinking. Green cells represent cases where thinking improved performance and red cells where it impaired performance. We see that (a)  only models above a certain size benefit from thinking; (b) for GP structures (left side) thinking often does not improve performance. }
\label{tab:thinking_relative_change}
\end{table*}

\subsection{Human procedure}

Native English speakers were recruited via the Prolific platform.\footnote{\url{https://www.prolific.com/}}  Sentences were displayed word-by-word, with each word shown for 400ms and a 100ms blank screen between words. After the sentence, the comprehension question was presented for 5 seconds. If unanswered within this time, the response was marked as incorrect. Participants completed two practice items, followed by one experimental sentence and one question, i.e., each participant saw a single experimental item and answered a single experimental question. This single-trial design has been shown to prevent fatigue \cite{christianson2022if} and learning effects \cite{fine2013rapid}. Each of the sentence-question pairs was shown to 10 participants, for a total of 5380 data points. The average completion time was 1:42 minutes, and participants were compensated with 0.30£ equivalent to 10.58£ per hour. 
The experiment was approved by the Ethics Committee at Tel-Aviv University. 

\subsection{LLM procedure}

We used few-shot prompting with LLMs, where each example includes a sentence, a question, and the correct answer. The examples did not contain any of our structures, to prevent in-context learning. Each model was prompted 8 times, using two system prompts and four example orderings. See Appendix \ref{app:prompt_ise} for an example of a prompt with each system prompt. For thinking models, we measured the performance both when thinking mode is on and off.\footnote{OpenAI models do not allow turning off thinking, so we set the thinking effort to low and high accordingly.} Due to the high resources required when thinking is allowed, in the thinking setup, we prompted the model only once.
We extract the probabilities of the correct and incorrect answer tokens, averaging these across the 8 prompts when needed. 

We test models from different families, sizes (31 models in total):
\begin{enumerate}[leftmargin=*,nosep]
    \item GPT family \cite{gpt4}: \emph{GPT-4o}, \emph{GPT-4.1}, \emph{o3}, \emph{o3-mini}, \emph{o4-mini}, \emph{GPT-5}.
    \item Llama-3 \cite{llama3herdmodels}: All models from the Llama-3.2 family on HuggingFace \cite{huggingface}.\footnote{\url{https://huggingface.co/models}}
    \item Qwen-3 \cite{qwen3}: All Qwen-3 models on HuggingFace.
    \item Gemma-3 \cite{gemma3}: All Gemma-3 models on HuggingFace.
    \item Distilled Qwen DeepSeek \cite{guo2025deepseek}: The DeepSeek R1 Qwen distilled models (except for the 32B version).
\end{enumerate}

\subsection{High-level results}
\label{subsec:results}

Table \ref{tab:model_accuracy}  presents average performance of humans and of the 3 best models per family on the target condition of each structure. For the results of the remaining models, see Appendix \ref{app:full_res}. 

\paragraph{Human performance} Human average performance across tasks is 28.3\%, confirming that the sentences used are indeed challenging for humans. Looking at the different structures, we see that for humans there is no single structure that is significantly harder than the other. GP accuracy ranges from 13.3\%-41.7\%, and the accuracy of the remaining structures ranges from 28.0\% (depth charge) to 36.9 \% (interference). This is unlike language models, as we analyze below.

\pgfplotstableread[row sep=\\,col sep=&]{
    Model & GP average & Depth charge & Double center & Interference \\
    Gemma-27B & 0.104 & 0.100 & 0.057 & 0.099 \\
    Gemma-27B-Instruct & 0.101 & 0.250 & 0.182 & 0.451 \\
    Llama-90B & 0.092 & 0.150 & 0.557 & 0.561 \\
    Llama-90B-Instruct & 0.084 & 0.150 & 0.578 & 0.611 \\
    Deepseek-14B & 0.119 & 0.460 & 0.338 & 0.281 \\
    Deepseek-14B - Thinking & 0.0996 & 0.24 & 0.5575 & 0.631 \\
    Qwen-32B & 0.107 & 0.240 & 0.367 & 0.501 \\ 
    Qwen-32B - Thinking & 0.223 & 0.430 & 0.5575 & 0.6313 \\ 
    GPT-5 & 0.209 & 0.55 & 0.658 & 0.631 \\
    GPT-5 - Thinking & 0.6267 & 0.54 & 0.658 & 0.6213 \\
    }\accuracydiff

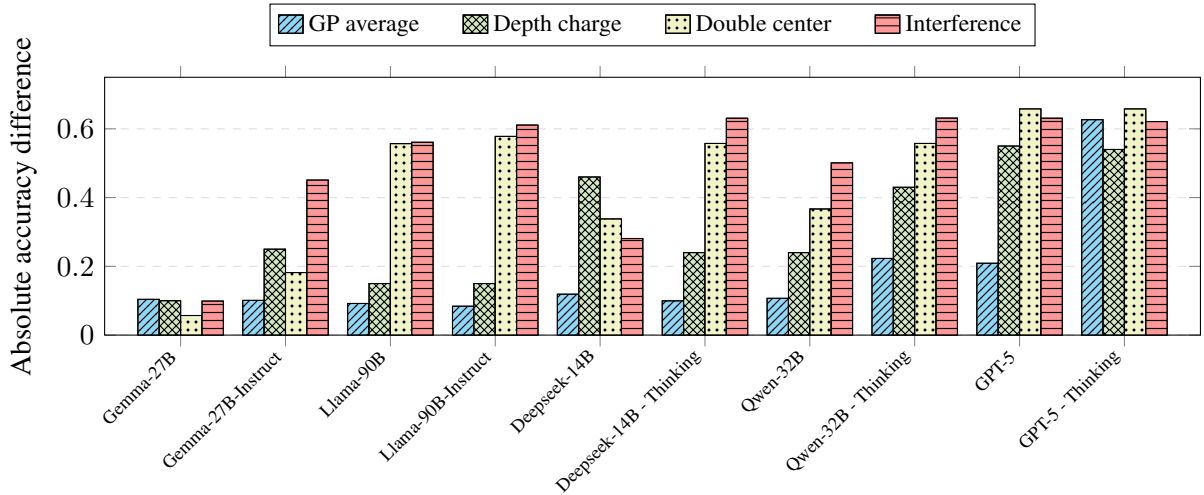
\begin{figure*}[t!]
\centering
\begin{tikzpicture}
\def\BarW{8pt}
\begin{axis}[
  ybar,
  bar width=\BarW,
  width=16cm,
  height=5cm,
  enlarge x limits={abs=1cm}, 
  ymin=0, ymax=0.75,
  ylabel={Absolute accuracy difference},
  ymajorgrids,
  grid style={dashed,gray!25},
  symbolic x coords={
    Gemma-27B,
    Gemma-27B-Instruct,
    Llama-90B,
    Llama-90B-Instruct,
    Deepseek-14B,
    Deepseek-14B - Thinking,
    Qwen-32B,
    Qwen-32B - Thinking,
    GPT-5,
    GPT-5 - Thinking
  },
  xtick=data,
  x tick label style={rotate=45, anchor=east, font=\scriptsize},
  legend image code/.code={
    \path[draw=black, fill=#1] (0cm,-0.11cm) rectangle (0.32cm,0.13cm);
    \path[pattern=\pgfkeysvalueof{/pgfplots/legend pattern},
          pattern color=black] (0cm,-0.11cm) rectangle (0.32cm,0.13cm);
  },
  legend style={
    at={(0.5,1.12)},
    anchor=south,
    legend columns=-1,
    /tikz/every even column/.append style={column sep=0.5cm},
    font=\small
  }
]
\pgfplotsset{
  /pgfplots/legend pattern/.initial=,
  series A/.style={fill=cyan!45,           /pgfplots/legend pattern=north east lines},
  series B/.style={fill=YellowGreen!45,    /pgfplots/legend pattern=crosshatch},
  series C/.style={fill=GreenYellow!55,    /pgfplots/legend pattern=dots},
  series D/.style={fill=Thistle!60,        /pgfplots/legend pattern=horizontal lines},
}
\addplot+[
    ybar, bar shift=-1.5*\BarW,
    draw=black,
    fill=cyan!40,            
     postaction={
        pattern=north east lines,
        pattern color=black,
    }
] table[x=Model, y={GP average}] {\accuracydiff};
\addlegendentry{GP average}

\addplot+[
    ybar, bar shift=-0.5*\BarW,
    draw=black,
    fill=YellowGreen!40,
    postaction={
        pattern=crosshatch,
        pattern color=black,
    }
] table[x=Model, y={Depth charge}] {\accuracydiff};
\addlegendentry{Depth charge}

\addplot+[
    ybar, bar shift=0.5*\BarW,
    draw=black,
    fill=GreenYellow!40,
    postaction={
        pattern=dots,
        pattern color=black,
    }
] table[x=Model, y={Double center}] {\accuracydiff};
\addlegendentry{Double center}

\addplot+[
    ybar, bar shift=1.5*\BarW,
    draw=black,
    fill=red!40,
    postaction={
        pattern=horizontal lines,
        pattern color=black,
    }
] table[x=Model, y={Interference}] {\accuracydiff};
\addlegendentry{Interference}
\end{axis}
\end{tikzpicture}
\caption{Absolute difference between LLM and human accuracy on the difficult condition.}
\label{fig:accuracy_diff}
\end{figure*}

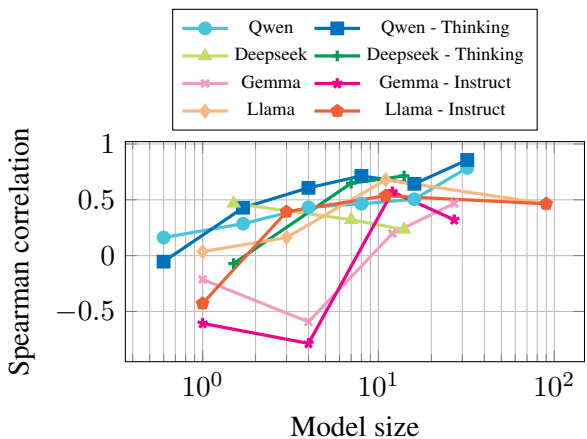
\begin{figure}[t!]
    \centering
\begin{tikzpicture}
\begin{axis}[
    xlabel={Model size},
    ylabel={Spearman correlation},
    xmode=log,
    log basis x=10,
    width = \columnwidth*0.99,
    height = 4.5cm,    
    legend style={
    at={(0.5,1.05)},
    anchor=south,
    legend columns=2,
    font=\scriptsize},
    cycle list name=color list,
    mark size=2pt,
    grid=both
]

\addplot[very thick,color=SkyBlue,mark=*] coordinates {(0.6,0.162) (1.7,0.286) (4,0.432) (8,0.464) (16,0.505) (32,0.786)};
\addlegendentry{Qwen}

\addplot[very thick,color=RoyalBlue,mark=square*] coordinates {(0.6,-0.054) (1.7,0.429) (4,0.607) (8,0.714) (16,0.643) (32,0.857)};
\addlegendentry{Qwen - Thinking}

\addplot[very thick,color=SpringGreen,mark=triangle*] coordinates {(1.5,0.468) (7,0.321) (14,0.234)};
\addlegendentry{Deepseek}

\addplot[very thick,color=ForestGreen,mark=+] coordinates {(1.5,-0.071) (7,0.649) (14,0.714)};
\addlegendentry{Deepseek - Thinking}

\addplot[very thick,color=Lavender,mark=x] coordinates {(1,-0.214) (4,-0.593) (12,0.198) (27,0.468)};
\addlegendentry{Gemma}

\addplot[very thick,color=Magenta,mark=star] coordinates {(1,-0.607) (4,-0.786) (12,0.571) (27,0.321)};
\addlegendentry{Gemma - Instruct}

\addplot[very thick,color=Apricot,mark=diamond] coordinates {(1,0.036) (3,0.162) (11,0.679) (90,0.464)};
\addlegendentry{Llama}

\addplot[very thick,color=RedOrange,mark=pentagon*] coordinates {(1,-0.429) (3,0.393) (11,0.536) (90,0.464)};
\addlegendentry{Llama - Instruct}

\end{axis}
\end{tikzpicture}
    \caption{Spearman correlation between humans and LLMs on ranking the difficulty of different structures.}
    \label{fig:spearman_correlation}
\end{figure}



\paragraph{LLM performance} While LLM performance is higher than humans', it is still far from perfect, with \emph{o3} achieving the best performance at 74.5\% without thinking and GPT-5 at 88.9\% with thinking. This shows that the structures are challenging for LLMs as well. When looking at each structure, an interesting finding appears: GP sentences are relatively harder for LLMs than the other structures, This trend is particularly noticeable for OpenAI models, and missing from Llama models.

\paragraph{Influence of size} 
We do not see clear scaling or inverse scaling behavior on any of the structures. models from the Llama, Qwen and Deepseek families do show some scaling trend on the difficult conditions in Double center embedding and Interference structures, but Gemma does not. See Appendix \ref{app:full_res} for a  detailed breakdown.

\paragraph{Influence of thinking} 
Table \ref{tab:thinking_relative_change} shows the effect of thinking on LLM performance in the different  structures. 
First, thinking helps once a model has enough parameters -- for models that are too weak from the DeepSeek and Qwen families thinking impairs performance, but once they have enough parameters, thinking helps. 
Second, thinking helps more uniformly on non-GP structures, while in GP structures thinking more often reduces performance. A notable exception is GPT-5's large improvement from thinking on GP sentences.

Overall, LLMs show human-like difficulties in processing these structures, especially garden-path (GP) sentences.  
One conjecture for the difference between GP and non-GP structures is how they relate to working memory -- 
interference and double center-embedded sentences are demanding because they heavily tax working memory, while depth-charge sentences require logical reasoning and working memory resources. In contrast, GP sentences do not stress working memory but require discarding of a misinterpretation (see Section \ref{sec:data_used}). Since LLMs have a larger working memory, they do better on non-GP sentences.

\section{Similarity between humans and LLMs}

The fact that structures challenging for humans are challenging for LLMs does not imply that humans and LLMs struggle in the same \textbf{manner}. We now investigate the similarity between humans and LLMs to answer 3 questions:
\begin{itemize}[leftmargin=*,nosep]
    \item \emph{On which structures is LLM performance closer to humans on the target condition?}
    \item \emph{How similar is the ordering of difficulty of the target conditions across structures between humans and LLMs?}
    \item \emph{When does the difference in difficulty between baseline and target sentences appear?}
\end{itemize}

\subsection{Similarity between LLM and human absolute performance}

On what structures is LLM performance (on the target condition) similar to humans? Figure \ref{fig:accuracy_diff} shows accuracy differences between humans and the largest model from each family, with and without thinking, on each target condition. Results for the four GP types are macro-averaged; complete results are provided in Appendix \ref{app:diff_humans}.

Overall, LLM performance is closer to humans on GP sentences than on other structures. When looking at all 31 models, this is confirmed with an average absolute accuracy difference  $0.173$ for GP structures, $0.328$ for depth charge, $0.330$ for double center embedding and $0.37$ for interference. 
This aligns with our observation from Section \ref{subsec:results} that GP sentences are harder for strong LLMs; because they are harder, performance on them is closer to humans than on the other structures. 

Second, the difference in accuracy between humans and LLMs on double center embedding and interference sentences is significantly higher than for depth charge and GP sentences. However, this is true \emph{only} for the larger models, since when averaged over all models, the difference in average absolute accuracy of depth charge sentences is close to double center embedding and interference sentences. For these stronger models, the absolute difference from humans is higher because they \emph{outperform} humans.

\subsection{Ranking of structure difficulty}

In Figure \ref{fig:spearman_correlation}, we see the Spearman rank correlation between different models and humans for the open-weight models, where we rank the seven structures by average performance. Contrary to absolute performance, a scaling trend emerges, with larger models having higher correlation to humans (except for Deepseek  without thinking). Additionally, not only does correlation grow, but it always eventually becomes positive.

The model with highest Spearman rank correlation is \emph{o4-mini} (not shown, due to its closed weights) with a correlation of 0.929. The open-weights model with highest Spearman rank correlation (0.857) is \emph{Qwen-32B-thinking}.

The main cause of difference in difficulty ranking between humans and LLMs is the fact that GP sentences are harder for models than the other structures. The Spearman rank correlation on GP structures alone or on non-GP structures alone is higher than on all structures together.

\subsection{Baseline vs. target conditions}

We say that humans and LLMs agree on the difficulty \emph{direction} in a structure if the LLM accuracy on the baseline condition is higher than on the target condition, like humans. The relative difficulty of target vs. baseline sentences is the central observation used to develop models of human cognition. Thus, it is important to verify that LLMs and humans share the same direction for each of our structures. We found that LLMs in general have the same \emph{directionality} as humans, i.e., the accuracy on the target condition is lower than the accuracy on the baseline condition. However, there are two cases where this does not hold: when the model is too weak, and therefore both conditions are equally bad, or when the model is too strong, and both conditions are equally good. We say that when an LLM is strong enough but not too strong, it is in the ``sweet spot''. How strong a model needs to be in order to be in the sweet spot for a structure depends on how hard the structure is for humans and on the type of structure (GP or non-GP). LLMs get stronger when they have more parameters or when thinking is on. 
Our findings echo the rule of \citet{2b_tokens_training}, which identified models up to 2B parameters as optimal for predicting human reading times. We refine this rule by showing that the cutoff varies by structure and that models below 2B parameters are not consistently optimal.

We can now define what is a \emph{violation} of our \emph{sweet spot rule}. A model B represents a violation of the sweet spot rule if a smaller model A and a larger model C both have the same directionality as humans, but the model B does not, where models A, B, and C are from the same family. 

All model  families present similar trends, so we present results for only 3 families with and without thinking. Full results in Appendix \ref{app:full_direc}.

\begin{table}[t!]
\centering
\scriptsize
\renewcommand{\arraystretch}{1}
\begin{tabular}{l|c|c|c|c}
\toprule
Model & Subj/Obj & NP/VP & NP/S & Red. relative\\
\hline
Llama-1B & \cmark & \xmark & \xmark & \cmark \\
Llama-1B-Instruct & \cmark & \xmark & \xmark & \cmark \\
Llama-3B & \cmark & \xmark & \xmark & \cmark \\ 
Llama-3B-Instruct & \cmark & \xmark & \xmark & \cmark \\
Llama-11B & \xmark & \cmark & \xmark & \cmark \\
Llama-11B-Instruct & \xmark & \cmark & \xmark & \cmark \\
Llama-90B & \xmark & \cmark & \cmark & \cmark \\
Llama-90B-Instruct & \xmark & \cmark & \cmark & \cmark \\\hline
Qwen-0.6B & \xmark & \xmark & \xmark & \cmark \\ 
Qwen-1.7B & \cmark & \cmark & \xmark & \cmark \\ 
Qwen-4B & \cmark & \cmark & \cmark & \cmark \\ 
Qwen-8B & \xmark & \cmark & \xmark & \cmark \\
Qwen-14B & \xmark & \cmark & \cmark & \cmark \\
Qwen-32B & \cmark & \cmark & \cmark & \cmark \\ \hdashline
Qwen-0.6B - thinking & \cmark & \cmark & \xmark & \cmark \\
Qwen-1.7B - thinking & \cmark & \cmark & \cmark & \cmark \\ 
Qwen-4B - thinking & \cmark & \cmark & \cmark & \cmark \\ 
Qwen-8B - thinking & \cmark & \cmark & \cmark & \cmark \\
Qwen-14B - thinking & \cmark & \cmark & \cmark & \cmark \\
Qwen-32B - thinking & \cmark & \cmark & \cmark & \cmark \\ \hline
GPT-4o & \xmark & \cmark & \cmark & \cmark \\
GPT-4.1 & \xmark & \cmark & \cmark & \cmark \\
o3-mini & \cmark & \cmark & \cmark & \cmark \\
o4-mini & \cmark & \cmark & \cmark & \cmark \\
o3 & \xmark & \cmark & \cmark & \cmark \\ 
GPT-5 & \xmark & \cmark & \cmark & \cmark \\ \hdashline
o3-mini - Thinking & \cmark & \cmark & \cmark & \cmark \\
o4-mini - Thinking & \cmark & \cmark & \cmark & \cmark \\
o3 - Thinking & \xmark & \cmark & \cmark & \xmark \\
GP-5 - Thinking &  \xmark & \cmark & \cmark & \cmark \\ \hline
\bottomrule
\end{tabular}
\caption{Target vs. baseline sentences. Crosses mark cases where the accuracy on the baseline is lower than on the target , and check signs mark the opposite.}
\label{tab:remaining_complex}
\end{table}

Table \ref{tab:remaining_complex} shows for each GP structure whether performance on the baseline structure is higher than the target structure. Table \ref{tab:gp_complex} shows the same data for the remaining structures. In both tables, checkmarks represent cases in which directionality is respected, crosses cases in which it is not.

\paragraph{Subj/Obj GP} For Subject/Object, OpenAI models get gradually better on the target condition, up to a point where performance on the target and baseline conditions is similar for o3-thinking. For Llama, the 1B and 3B have the directionality of humans, but the 11B and 90B do not. For the Qwen family, we have two violdations of our general rule, for sizes 8B and 14B, since they do not have the same directionality as humans, but their thinking counterparts are similar to humans. 

\paragraph{NP/VP GP} For the OpenAI family, all models have the same directionality as humans. For Llama, models of size larger than 3B behave like humans, and for Qwen only the 0.6B model is not strong enough. There are no violations to our general rule.

\paragraph{NP/S GP} While NP/S sentences are harder for humans than NP/VP, it is the opposite for LLMs. Therefore, models need  more parameters to be in the sweet spot. All models from OpenAI indeed have the same directionality as humans. For Llama models,  models of size larger than 11B or that are of size 11B and instruction-tuned are behave like humans. For Qwen, models of size smaller than 1.7B or 1.7B without thinking are not strong enough, while the rest are in the sweet spot, with Qwen-8B being a violation of our rule.

\paragraph{Reduced relative GP} All models from the Qwen and Llama families are strong enough to be similar to humans. The \emph{o3 + thinking} model from OpenAI is too good on the target condition and therefore not similar to humans.

\begin{table}[t!]
\centering
\scriptsize
\renewcommand{\arraystretch}{1}
\begin{tabular}{l|c|c|c}
\toprule
Model & Depth charge & Double center & Interference \\
\hline
Llama-1B & \xmark & \cmark & \xmark \\
Llama-1B-Instruct & \xmark & \cmark & \cmark  \\
Llama-3B & \cmark & \cmark & \cmark \\ 
Llama-3B-Instruct & \cmark & \cmark & \cmark \\
Llama-11B & \cmark & \cmark & \cmark \\
Llama-11B-Instruct & \cmark & \cmark & \cmark \\
Llama-90B & \cmark & \xmark & \xmark \\
Llama-90B-Instruct & \cmark & \xmark & \xmark \\\hline
Qwen-0.6B & \xmark & \cmark & \xmark  \\ 
Qwen-1.7B & \xmark & \cmark & \xmark \\ 
Qwen-4B & \cmark & \cmark & \cmark \\ 
Qwen-8B & \cmark & \cmark & \cmark \\
Qwen-14B & \cmark & \cmark & \xmark \\
Qwen-32B & \cmark & \cmark & \xmark \\ \hdashline
Qwen-0.6B - Thinking & \cmark & \cmark & \xmark \\
Qwen-1.7B - Thinking & \cmark & \cmark & \xmark \\ 
Qwen-4B - Thinking & \cmark & \cmark & \xmark \\ 
Qwen-8B - Thinking & \cmark & \cmark & \xmark \\
Qwen-14B - Thinking & \cmark & \xmark & \xmark \\
Qwen-32B - Thinking & \cmark & \xmark & \xmark \\ \hline
GPT-4o & \cmark & \xmark & \cmark  \\
GPT-4.1 & \cmark & \xmark & \xmark  \\
o3-mini & \cmark & \xmark & \xmark \\ 
o4-mini & \cmark & \cmark & \xmark \\
o3 & \cmark & \xmark & \xmark  \\
GPT-5 & \cmark & \xmark & \xmark \\ \hdashline
o3-mini - Thinking & \cmark & \cmark & \xmark \\
o4-mini - Thinking & \cmark & \cmark & \xmark  \\
o3 - Thinking & \cmark & \xmark & \xmark  \\
GPT-5 - Thinking & \cmark & \xmark & \xmark \\ \hline
\bottomrule
\end{tabular}
\caption{Target vs. baseline for non GP structures.}
\label{tab:gp_complex}
\end{table}

\paragraph{Depth charge} All models from the OpenAI family are strong enough to be similar to humans. Models under 2B without thinking are not strong enough for Llama and Qwen.

\paragraph{Double center} For the OpenAI family, o3-mini with thinking and o4-mini with or without thinking keep directionality, while other models show comparable accuracy on the target and baseline conditions. For the Llama family, only the 90B models are too strong to be similar to humans. Only the two largest Qwen models in thinking mode are too strong to be similar to humans.

\paragraph{Interference} All OpenAI models but GPT-4o are too strong and perform equally well on the target and baseline sentences. For Qwen, the 0.6B and 1.7B models without thinking are too weak to be similar to humans, the 4B and 8B models are in the sweet spot, and the larger models and all the thinking models are too strong. For Llama, only the 90B models are too strong to be similar to humans, and the 1B model is not strong enough.


\pgfplotstableread[row sep=\\,col sep=&]{
    Syntactic structure & Violation rate \\
    Subj/Obj & 0.143  \\
    NP/VP & 0.0  \\
    Depth charge & 0.036 \\
    NP/S & 0.036 \\
    Reduced relative & 0.071 \\
    Interference & 0.036 \\
    Double center & 0.0 \\
    }\violationrate

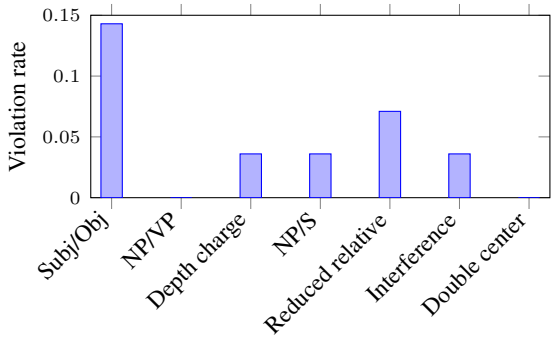
\begin{figure}[t!]
  \centering
\begin{tikzpicture}[trim left=0pt]
    \begin{axis}[
            ybar,
            width = \columnwidth*0.99,
            height = 4cm,
            symbolic x coords={Subj/Obj,NP/VP,Depth charge,NP/S,Reduced relative,Interference,Double center},
            xtick=data,
            tick label style={font=\small}, 
            label style={font=\small},      
            x tick label style={
                rotate=45,                  
                anchor=east,                
                font=\small,           
            },
            bar width=8pt,                  
            ymin=0, 
            ymax=0.15,
            enlarge x limits=0.05,         
            ylabel={Violation rate},        
            y tick label style={
                font=\scriptsize,
                /pgf/number format/fixed,
                /pgf/number format/precision=3
            },
            trim axis left,
            enlarge y limits=false,
            y label style={
                at={(axis description cs:-0.12,0.5)}, 
                anchor=south,
            },
        ]
        \addplot table[x=Syntactic structure,y=Violation rate]{\violationrate};
    \end{axis}
\end{tikzpicture}
  \caption{Violation rate per structure}
  \label{fig:transgression_rate}
\end{figure}


\paragraph{Violation rate} To evaluate the sweet spot rule, we measure for each structure the \emph{violation rate}. The violation rate is the total number of models that represent a violation divided by the total number of models that could represent a violation (in our case 28).\footnote{We do not include OpenAI models because their size is unknown, and thus ordering them is not possible.} Figure \ref{fig:transgression_rate} shows the violation rate per structure. All structures but Subj/Obj have an extremely low violation rate, with two violating models at most. For Subj/Obj, we conjecture that the high violation rate is due to the relatively small difference in performance between the target and baseline conditions in humans (0.133 vs. 0.191).

\paragraph{Influence of thinking} Thinking influences directionality insofar as it makes the model stronger. For GP structures, almost all thinking models have the same directionality as humans. For the other structures, all thinking models have the same directionality as humans for the hardest structure (depth charge) and all do not have the same directionality on the easiest structure (interference). For double center, some thinking models have the same directionality as humans but some do not.
\section{Conclusion}

This study explores the similarity between human and LLM sentence processing. Using 7 structures whose difficulty has been studied in previous psycholinguistic works, we study whether LLMs and humans make similar comprehension errors. Our findings demonstrate that
LLMs behave differently on GP structures compared to other structures -- LLMs have low performance on our structures, like humans, but as LLMs become stronger they perform better than humans on non GP structures. We conjecture that this difference between GP and non-GP sentences is related to the advantage LLMs have on humans in terms of working memory and their ability to perform logical reasoning. Designing and running experiments to test this conjecture is a key direction for future work.

In addition, we find that LLM errors are similar to humans' when they are not too strong and not too weak (the ``sweet-spot'' rule): indeed in this case LLMs perform much better on the baseline condition compared to the target condition, while if they are too weak (or too strong) they perform equally poorly (or well) on both conditions.

Overall, our paper offers multiple new insights on the similarity between humans and LLMs sentence processing mechanisms, which can be further studied in future work.
\section*{Limitations}

In this study, we evaluated reading comprehension across a wide array of LLMs. However, we studied only one family of closed-source models and did not test models from Anthropic or Google. Measuring their understanding can be interesting for future works. Additionally, we did not try every type GP. Measuring performance on additional GP types would validate the difference we found between GP and non GP sentences.r Finally, we did not collect data on metrics beyond reading comprehension, such as eye gaze or reading time,
Gathering such metrics and analyzing their correlation with sentence comprehension could provide valuable insights.

\bibliography{custom}

\begin{thebibliography}{43}
\providecommand{\natexlab}[1]{#1}

\bibitem[{Almazrouei et~al.(2023)Almazrouei, Alobeidli, Alshamsi, Cappelli, Cojocaru, Debbah, Étienne Goffinet, Hesslow, Launay, Malartic, Mazzotta, Noune, Pannier, and Penedo}]{falcon_series}
Ebtesam Almazrouei, Hamza Alobeidli, Abdulaziz Alshamsi, Alessandro Cappelli, Ruxandra Cojocaru, Mérouane Debbah, Étienne Goffinet, Daniel Hesslow, Julien Launay, Quentin Malartic, Daniele Mazzotta, Badreddine Noune, Baptiste Pannier, and Guilherme Penedo. 2023.
\newblock \href {https://arxiv.org/abs/2311.16867} {The falcon series of open language models}.

\bibitem[{Amouyal et~al.(2024)Amouyal, Meltzer-Asscher, and Berant}]{pretesting}
Samuel Amouyal, Aya Meltzer-Asscher, and Jonathan Berant. 2024.
\newblock \href {https://aclanthology.org/2024.findings-eacl.12} {Large language models for psycholinguistic plausibility pretesting}.
\newblock In \emph{Findings of the Association for Computational Linguistics: EACL 2024}, pages 166--181, St. Julian{'}s, Malta. Association for Computational Linguistics.

\bibitem[{Amouyal et~al.(2025)Amouyal, Meltzer-Asscher, and Berant}]{amouyal-etal-2025-lm}
Samuel~Joseph Amouyal, Aya Meltzer-Asscher, and Jonathan Berant. 2025.
\newblock \href {https://doi.org/10.18653/v1/2025.acl-long.403} {When the {LM} misunderstood the human chuckled: Analyzing garden path effects in humans and language models}.
\newblock In \emph{Proceedings of the 63rd Annual Meeting of the Association for Computational Linguistics (Volume 1: Long Papers)}, pages 8235--8253, Vienna, Austria. Association for Computational Linguistics.

\bibitem[{Cacheteux and King(2022)}]{cacheteux-middle-layer}
Charlotte Cacheteux and Jean-Rémi King. 2022.
\newblock \href {https://doi.org/10.1038/s42003-022-03036-1} {An activation-based model of sentence processing as skilled memory retrieval}.
\newblock \emph{Nature}, pages 375--419.

\bibitem[{Christianson et~al.(2022)Christianson, Dempsey, Tsiola, and Goldshtein}]{christianson2022if}
Kiel Christianson, Jack Dempsey, Anna Tsiola, and Maria Goldshtein. 2022.
\newblock What if they're just not that into you (or your experiment)? on motivation and psycholinguistics.
\newblock In \emph{Psychology of learning and motivation}, volume~76, pages 51--88. Elsevier.

\bibitem[{Christianson et~al.(2001)Christianson, Hollingworth, Halliwell, and Ferreira}]{christianson2001}
Kiel Christianson, Andrew Hollingworth, John~F Halliwell, and Fernanda Ferreira. 2001.
\newblock Thematic roles assigned along the garden path linger.
\newblock \emph{Cognitive psychology}, 42(4):368--407.

\bibitem[{Christianson et~al.(2006)Christianson, Williams, Zacks, and Ferreira}]{christianson2006}
Kiel Christianson, Carrick~C Williams, Rose~T Zacks, and Fernanda Ferreira. 2006.
\newblock Misinterpretations of garden-path sentences by older and younger adults.
\newblock \emph{Discourse Processes}, 42:205--238.

\bibitem[{Ferreira and Henderson(1990)}]{gp3}
Fernanda Ferreira and John~M Henderson. 1990.
\newblock Use of verb information in syntactic parsing: evidence from eye movements and word-by-word self-paced reading.
\newblock \emph{Journal of Experimental Psychology: Learning, Memory, and Cognition}, 16(4):555.

\bibitem[{Ferreira and Yang(2019)}]{ferreira2019problem}
Fernanda Ferreira and Zoe Yang. 2019.
\newblock The problem of comprehension in psycholinguistics.
\newblock \emph{Discourse Processes}, 56(7):485--495.

\bibitem[{Fine et~al.(2013)Fine, Jaeger, Farmer, and Qian}]{fine2013rapid}
Alex~B Fine, T~Florian Jaeger, Thomas~A Farmer, and Ting Qian. 2013.
\newblock Rapid expectation adaptation during syntactic comprehension.
\newblock \emph{PloS one}, 8(10):e77661.

\bibitem[{Frazier(1987)}]{frazier_newer}
Lyn Frazier. 1987.
\newblock Sentence processing: A tutorial review.

\bibitem[{Garnsey et~al.(1997)Garnsey, Pearlmutter, Myers, and Lotocky}]{gp1}
Susan~M Garnsey, Neal~J Pearlmutter, Elizabeth Myers, and Melanie~A Lotocky. 1997.
\newblock The contributions of verb bias and plausibility to the comprehension of temporarily ambiguous sentences.
\newblock \emph{Journal of memory and language}, 37(1):58--93.

\bibitem[{Gemini(2024)}]{geminiteam2024gemini}
Team Gemini. 2024.
\newblock \href {https://arxiv.org/abs/2403.05530} {Gemini 1.5: Unlocking multimodal understanding across millions of tokens of context}.

\bibitem[{Gibson and Thomas(1999)}]{gibson_interference}
Edward Gibson and James Thomas. 1999.
\newblock Memory limitations and structural forgetting: The perception of complex ungrammatical sentences as grammatical.
\newblock \emph{Language and cognitive processes}, 14(3):225--248.

\bibitem[{Goldstein et~al.(2023)Goldstein, Ham, Nastase, Zada, Grinstein-Dabus, Aubrey, Schain, Gazula, Feder, Doyle, Devore, Dugan, Friedman, Brenner, Hassidim, Devinsky, Flinker, Levy, and Hasson}]{Goldstein2023CorrespondenceBT}
Ariel Goldstein, Eric Ham, Samuel~A. Nastase, Zaid Zada, Avigail Grinstein-Dabus, Bobbi Aubrey, Mariano Schain, Harshvardhan Gazula, Amir Feder, Werner~K. Doyle, Sasha Devore, Patricia Dugan, Daniel Friedman, Michael~P. Brenner, Avinatan Hassidim, Orrin Devinsky, Adeen Flinker, Omer Levy, and Uri Hasson. 2023.
\newblock \href {https://api.semanticscholar.org/CorpusID:250534035} {Correspondence between the layered structure of deep language models and temporal structure of natural language processing in the human brain}.
\newblock \emph{bioRxiv}.

\bibitem[{Gordon et~al.(2001)Gordon, Hendrick, and Johnson}]{memory-interference}
Peter~C. Gordon, Randall Hendrick, and M.~K. Johnson. 2001.
\newblock Memory interference during language processing.
\newblock \emph{Journal of experimental psychology. Learning, memory, and cognition}, 27 6:1411--23.

\bibitem[{Grattafiori et~al.(2024)Grattafiori, Dubey, Jauhri, Pandey, Kadian, Al-Dahle, Letman, Mathur, Schelten, Vaughan, Yang, Fan, Goyal, Hartshorn, Yang, Mitra, Sravankumar, Korenev, Hinsvark, Rao, Zhang, Rodriguez, Gregerson, Spataru, Roziere, Biron, Tang, Chern, Caucheteux, Nayak, Bi, Marra, McConnell, Keller, Touret, Wu, Wong, Ferrer, Nikolaidis, Allonsius, Song, Pintz, Livshits, Wyatt, Esiobu, Choudhary, Mahajan, Garcia-Olano, Perino, Hupkes, Lakomkin, AlBadawy, Lobanova, Dinan, Smith, Radenovic, Guzmán, Zhang, Synnaeve, Lee, Anderson, Thattai, Nail, Mialon, Pang, Cucurell, Nguyen, Korevaar, Xu, Touvron, Zarov, Ibarra, Kloumann, Misra, Evtimov, Zhang, Copet, Lee, Geffert, Vranes, Park, Mahadeokar, Shah, van~der Linde, Billock, Hong, Lee, Fu, Chi, Huang, Liu, Wang, Yu, Bitton, Spisak, Park, Rocca, Johnstun, Saxe, Jia, Alwala, Prasad, Upasani, Plawiak, Li, Heafield, Stone, El-Arini, Iyer, Malik, Chiu, Bhalla, Lakhotia, Rantala-Yeary, van~der Maaten, Chen, Tan, Jenkins, Martin, Madaan, Malo, Blecher,
  Landzaat, de~Oliveira, Muzzi, Pasupuleti, Singh, Paluri, Kardas, Tsimpoukelli, Oldham, Rita, Pavlova, Kambadur, Lewis, Si, Singh, Hassan, Goyal, Torabi, Bashlykov, Bogoychev, Chatterji, Zhang, Duchenne, Çelebi, Alrassy, Zhang, Li, Vasic, Weng, Bhargava, Dubal, Krishnan, Koura, Xu, He, Dong, Srinivasan, Ganapathy, Calderer, Cabral, Stojnic, Raileanu, Maheswari, Girdhar, Patel, Sauvestre, Polidoro, Sumbaly, Taylor, Silva, Hou, Wang, Hosseini, Chennabasappa, Singh, Bell, Kim, Edunov, Nie, Narang, Raparthy, Shen, Wan, Bhosale, Zhang, Vandenhende, Batra, Whitman, Sootla, Collot, Gururangan, Borodinsky, Herman, Fowler, Sheasha, Georgiou, Scialom, Speckbacher, Mihaylov, Xiao, Karn, Goswami, Gupta, Ramanathan, Kerkez, Gonguet, Do, Vogeti, Albiero, Petrovic, Chu, Xiong, Fu, Meers, Martinet, Wang, Wang, Tan, Xia, Xie, Jia, Wang, Goldschlag, Gaur, Babaei, Wen, Song, Zhang, Li, Mao, Coudert, Yan, Chen, Papakipos, Singh, Srivastava, Jain, Kelsey, Shajnfeld, Gangidi, Victoria, Goldstand, Menon, Sharma, Boesenberg,
  Baevski, Feinstein, Kallet, Sangani, Teo, Yunus, Lupu, Alvarado, Caples, Gu, Ho, Poulton, Ryan, Ramchandani, Dong, Franco, Goyal, Saraf, Chowdhury, Gabriel, Bharambe, Eisenman, Yazdan, James, Maurer, Leonhardi, Huang, Loyd, Paola, Paranjape, Liu, Wu, Ni, Hancock, Wasti, Spence, Stojkovic, Gamido, Montalvo, Parker, Burton, Mejia, Liu, Wang, Kim, Zhou, Hu, Chu, Cai, Tindal, Feichtenhofer, Gao, Civin, Beaty, Kreymer, Li, Adkins, Xu, Testuggine, David, Parikh, Liskovich, Foss, Wang, Le, Holland, Dowling, Jamil, Montgomery, Presani, Hahn, Wood, Le, Brinkman, Arcaute, Dunbar, Smothers, Sun, Kreuk, Tian, Kokkinos, Ozgenel, Caggioni, Kanayet, Seide, Florez, Schwarz, Badeer, Swee, Halpern, Herman, Sizov, Guangyi, Zhang, Lakshminarayanan, Inan, Shojanazeri, Zou, Wang, Zha, Habeeb, Rudolph, Suk, Aspegren, Goldman, Zhan, Damlaj, Molybog, Tufanov, Leontiadis, Veliche, Gat, Weissman, Geboski, Kohli, Lam, Asher, Gaya, Marcus, Tang, Chan, Zhen, Reizenstein, Teboul, Zhong, Jin, Yang, Cummings, Carvill, Shepard, McPhie,
  Torres, Ginsburg, Wang, Wu, U, Saxena, Khandelwal, Zand, Matosich, Veeraraghavan, Michelena, Li, Jagadeesh, Huang, Chawla, Huang, Chen, Garg, A, Silva, Bell, Zhang, Guo, Yu, Moshkovich, Wehrstedt, Khabsa, Avalani, Bhatt, Mankus, Hasson, Lennie, Reso, Groshev, Naumov, Lathi, Keneally, Liu, Seltzer, Valko, Restrepo, Patel, Vyatskov, Samvelyan, Clark, Macey, Wang, Hermoso, Metanat, Rastegari, Bansal, Santhanam, Parks, White, Bawa, Singhal, Egebo, Usunier, Mehta, Laptev, Dong, Cheng, Chernoguz, Hart, Salpekar, Kalinli, Kent, Parekh, Saab, Balaji, Rittner, Bontrager, Roux, Dollar, Zvyagina, Ratanchandani, Yuvraj, Liang, Alao, Rodriguez, Ayub, Murthy, Nayani, Mitra, Parthasarathy, Li, Hogan, Battey, Wang, Howes, Rinott, Mehta, Siby, Bondu, Datta, Chugh, Hunt, Dhillon, Sidorov, Pan, Mahajan, Verma, Yamamoto, Ramaswamy, Lindsay, Lindsay, Feng, Lin, Zha, Patil, Shankar, Zhang, Zhang, Wang, Agarwal, Sajuyigbe, Chintala, Max, Chen, Kehoe, Satterfield, Govindaprasad, Gupta, Deng, Cho, Virk, Subramanian, Choudhury,
  Goldman, Remez, Glaser, Best, Koehler, Robinson, Li, Zhang, Matthews, Chou, Shaked, Vontimitta, Ajayi, Montanez, Mohan, Kumar, Mangla, Ionescu, Poenaru, Mihailescu, Ivanov, Li, Wang, Jiang, Bouaziz, Constable, Tang, Wu, Wang, Wu, Gao, Kleinman, Chen, Hu, Jia, Qi, Li, Zhang, Zhang, Adi, Nam, Yu, Wang, Zhao, Hao, Qian, Li, He, Rait, DeVito, Rosnbrick, Wen, Yang, Zhao, and Ma}]{llama3herdmodels}
Aaron Grattafiori, Abhimanyu Dubey, Abhinav Jauhri, Abhinav Pandey, Abhishek Kadian, Ahmad Al-Dahle, Aiesha Letman, Akhil Mathur, Alan Schelten, Alex Vaughan, Amy Yang, Angela Fan, Anirudh Goyal, Anthony Hartshorn, Aobo Yang, Archi Mitra, Archie Sravankumar, Artem Korenev, Arthur Hinsvark, and 542 others. 2024.
\newblock \href {https://arxiv.org/abs/2407.21783} {The llama 3 herd of models}.

\bibitem[{Guo et~al.(2025)Guo, Yang, Zhang, Song, Zhang, Xu, Zhu, Ma, Wang, Bi et~al.}]{guo2025deepseek}
Daya Guo, Dejian Yang, Haowei Zhang, Junxiao Song, Ruoyu Zhang, Runxin Xu, Qihao Zhu, Shirong Ma, Peiyi Wang, Xiao Bi, and 1 others. 2025.
\newblock \href {https://arxiv.org/abs/2501.12948} {Deepseek-r1: Incentivizing reasoning capability in llms via reinforcement learning}.
\newblock \emph{ArXiv preprint}, abs/2501.12948.

\bibitem[{Hardt(2025)}]{sparks_double_center}
Daniel Hardt. 2025.
\newblock Sparks of pure competence in llms: the case of syntactic center embedding in english.
\newblock \emph{Society for Computation in Linguistics}, 8(1).

\bibitem[{Hu et~al.(2020)Hu, Gauthier, Qian, Wilcox, and Levy}]{hu-etal-2020-systematic}
Jennifer Hu, Jon Gauthier, Peng Qian, Ethan Wilcox, and Roger Levy. 2020.
\newblock \href {https://doi.org/10.18653/v1/2020.acl-main.158} {A systematic assessment of syntactic generalization in neural language models}.
\newblock In \emph{Proceedings of the 58th Annual Meeting of the Association for Computational Linguistics}, pages 1725--1744, Online. Association for Computational Linguistics.

\bibitem[{Hu et~al.(2024)Hu, Mahowald, Lupyan, Ivanova, and Levy}]{align_center_embedding}
Jennifer Hu, Kyle Mahowald, Gary Lupyan, Anna Ivanova, and Roger Levy. 2024.
\newblock Language models align with human judgments on key grammatical constructions.
\newblock \emph{Proceedings of the National Academy of Sciences}, 121(36):e2400917121.

\bibitem[{Irwin et~al.(2023)Irwin, Wilson, and Marantz}]{bert_gp}
Tovah Irwin, Kyra Wilson, and Alec Marantz. 2023.
\newblock \href {https://doi.org/10.18653/v1/2023.eacl-main.235} {{BERT} shows garden path effects}.
\newblock In \emph{Proceedings of the 17th Conference of the European Chapter of the Association for Computational Linguistics}, pages 3220--3232, Dubrovnik, Croatia. Association for Computational Linguistics.

\bibitem[{J{\"a}ger et~al.(2015)J{\"a}ger, Benz, Roeser, Dillon, and Vasishth}]{jager2015teasing}
Lena~A J{\"a}ger, Lena Benz, Jens Roeser, Brian~W Dillon, and Shravan Vasishth. 2015.
\newblock Teasing apart retrieval and encoding interference in the processing of anaphors.
\newblock \emph{Frontiers in psychology}, 6:506.

\bibitem[{Kizach et~al.(2016)Kizach, Christensen, and Weed}]{depth_charge}
Johannes Kizach, Ken~Ramsh{\o}j Christensen, and Ethan Weed. 2016.
\newblock A verbal illusion: now in three languages.
\newblock \emph{Journal of Psycholinguistic Research}, 45(3):753--768.

\bibitem[{Kuribayashi et~al.(2025)Kuribayashi, Oseki, Taieb, Inui, and Baldwin}]{kuribayashi2025large}
Tatsuki Kuribayashi, Yohei Oseki, Souhaib~Ben Taieb, Kentaro Inui, and Timothy Baldwin. 2025.
\newblock \href {https://arxiv.org/abs/2502.01615} {Large language models are human-like internally}.
\newblock \emph{ArXiv preprint}, abs/2502.01615.

\bibitem[{Linzen et~al.(2016)Linzen, Dupoux, and Goldberg}]{linzen-etal-2016-assessing}
Tal Linzen, Emmanuel Dupoux, and Yoav Goldberg. 2016.
\newblock \href {https://doi.org/10.1162/tacl_a_00115} {Assessing the ability of {LSTM}s to learn syntax-sensitive dependencies}.
\newblock \emph{Transactions of the Association for Computational Linguistics}, 4:521--535.

\bibitem[{Miller and Chomsky(1963)}]{double_center_chomsky}
George~A Miller and Noam Chomsky. 1963.
\newblock Finitary models of language users.

\bibitem[{Oh and Schuler(2023)}]{2b_tokens_training}
Byung-Doh Oh and William Schuler. 2023.
\newblock \href {https://doi.org/10.18653/v1/2023.findings-emnlp.128} {Transformer-based language model surprisal predicts human reading times best with about two billion training tokens}.
\newblock In \emph{Findings of the Association for Computational Linguistics: EMNLP 2023}, pages 1915--1921, Singapore. Association for Computational Linguistics.

\bibitem[{OpenAI(2023)}]{gpt4}
OpenAI. 2023.
\newblock \href {https://arxiv.org/abs/2303.08774} {Gpt-4 technical report}.
\newblock \emph{ArXiv preprint}, abs/2303.08774.

\bibitem[{Paape(2023)}]{paape2023role}
Dario Paape. 2023.
\newblock The role of incremental and superficial processing in the depth charge illusion: Experimental and modeling evidence.
\newblock \emph{Journal of Semantics}, 40(1):93--125.

\bibitem[{Rego et~al.(2024)Rego, Snell, and Meeter}]{Rego2024LanguageMO}
Adrielli~Lopes Rego, Joshua Snell, and Martijn Meeter. 2024.
\newblock \href {https://api.semanticscholar.org/CorpusID:269527476} {Language models outperform cloze predictability in a cognitive model of reading}.
\newblock \emph{PLOS Computational Biology}, 20.

\bibitem[{Ren et~al.(2024)Ren, Jin, Zhang, and Xiong}]{Ren2024DoLL}
Yuqi Ren, Renren Jin, Tongxuan Zhang, and Deyi Xiong. 2024.
\newblock \href {https://arxiv.org/abs/2402.18023} {Do large language models mirror cognitive language processing?}
\newblock \emph{ArXiv preprint}, abs/2402.18023.

\bibitem[{Saul et~al.(2025)Saul, Keshev, and Meltzer-Asscher}]{saul2025interference}
Niki Saul, Maayan Keshev, and Aya Meltzer-Asscher. 2025.
\newblock Interference in the formation of filler-gap dependencies: Evidence from hebrew relative clauses.
\newblock \emph{Journal of Memory and Language}, 143:104626.

\bibitem[{Schrimpf et~al.(2021)Schrimpf, Blank, Tuckute, Kauf, Hosseini, Kanwisher, Tenenbaum, and Fedorenko}]{fedorenko_brain_corr}
Martin Schrimpf, Idan~Asher Blank, Greta Tuckute, Carina Kauf, Eghbal~A. Hosseini, Nancy Kanwisher, Joshua~B. Tenenbaum, and Evelina Fedorenko. 2021.
\newblock \href {https://doi.org/10.1073/pnas.2105646118} {The neural architecture of language: Integrative modeling converges on predictive processing}.
\newblock \emph{Proceedings of the National Academy of Sciences}, 118(45):e2105646118.

\bibitem[{Sun and Wang(2024)}]{Sun2024ComputationalSM}
Kun Sun and Rong Wang. 2024.
\newblock \href {https://arxiv.org/abs/2403.15822} {Computational sentence-level metrics predicting human sentence comprehension}.
\newblock \emph{ArXiv preprint}, abs/2403.15822.

\bibitem[{Team et~al.(2025)Team, Kamath, Ferret, Pathak, Vieillard, Merhej, Perrin, Matejovicova, Ram{\'e}, Rivi{\`e}re et~al.}]{gemma3}
Gemma Team, Aishwarya Kamath, Johan Ferret, Shreya Pathak, Nino Vieillard, Ramona Merhej, Sarah Perrin, Tatiana Matejovicova, Alexandre Ram{\'e}, Morgane Rivi{\`e}re, and 1 others. 2025.
\newblock \href {https://arxiv.org/abs/2503.19786} {Gemma 3 technical report}.
\newblock \emph{ArXiv preprint}, abs/2503.19786.

\bibitem[{Trueswell et~al.(1993)Trueswell, Tanenhaus, and Kello}]{gp2}
John~C Trueswell, Michael~K Tanenhaus, and Christopher Kello. 1993.
\newblock Verb-specific constraints in sentence processing: separating effects of lexical preference from garden-paths.
\newblock \emph{Journal of Experimental psychology: Learning, memory, and Cognition}, 19(3):528.

\bibitem[{Vasishth et~al.(2010)Vasishth, Suckow, Lewis, and Kern}]{vasishth2010short}
Shravan Vasishth, Katja Suckow, Richard~L Lewis, and Sabine Kern. 2010.
\newblock Short-term forgetting in sentence comprehension: Crosslinguistic evidence from verb-final structures.
\newblock \emph{Language and Cognitive Processes}, 25(4):533--567.

\bibitem[{Villata et~al.(2018)Villata, Tabor, and Franck}]{villata2018encoding}
Sandra Villata, Whitney Tabor, and Julie Franck. 2018.
\newblock Encoding and retrieval interference in sentence comprehension: Evidence from agreement.
\newblock \emph{Frontiers in psychology}, 9:2.

\bibitem[{Warstadt et~al.(2019)Warstadt, Singh, and Bowman}]{acceptability_nn}
Alex Warstadt, Amanpreet Singh, and Samuel~R. Bowman. 2019.
\newblock \href {https://doi.org/10.1162/tacl_a_00290} {Neural network acceptability judgments}.
\newblock \emph{Transactions of the Association for Computational Linguistics}, 7:625--641.

\bibitem[{Wolf et~al.(2019)Wolf, Debut, Sanh, Chaumond, Delangue, Moi, Cistac, Rault, Louf, Funtowicz, and Brew}]{huggingface}
Thomas Wolf, Lysandre Debut, Victor Sanh, Julien Chaumond, Clement Delangue, Anthony Moi, Pierric Cistac, Tim Rault, R{\'e}mi Louf, Morgan Funtowicz, and Jamie Brew. 2019.
\newblock \href {https://arxiv.org/abs/1910.03771} {Huggingface's transformers: State-of-the-art natural language processing}.
\newblock \emph{ArXiv preprint}, abs/1910.03771.

\bibitem[{Yang et~al.(2025)Yang, Li, Yang, Zhang, Hui, Zheng, Yu, Gao, Huang, Lv, Zheng, Liu, Zhou, Huang, Hu, Ge, Wei, Lin, Tang, Yang, Tu, Zhang, Yang, Yang, Zhou, Zhou, Lin, Dang, Bao, Yang, Yu, Deng, Li, Xue, Li, Zhang, Wang, Zhu, Men, Gao, Liu, Luo, Li, Tang, Yin, Ren, Wang, Zhang, Ren, Fan, Su, Zhang, Zhang, Wan, Liu, Wang, Cui, Zhang, Zhou, and Qiu}]{qwen3}
An~Yang, Anfeng Li, Baosong Yang, Beichen Zhang, Binyuan Hui, Bo~Zheng, Bowen Yu, Chang Gao, Chengen Huang, Chenxu Lv, Chujie Zheng, Dayiheng Liu, Fan Zhou, Fei Huang, Feng Hu, Hao Ge, Haoran Wei, Huan Lin, Jialong Tang, and 41 others. 2025.
\newblock \href {https://arxiv.org/abs/2505.09388} {Qwen3 technical report}.

\bibitem[{Zhang et~al.(2023)Zhang, Ryskin, and Gibson}]{zhang2023noisy}
Yuhan Zhang, Rachel Ryskin, and Edward Gibson. 2023.
\newblock A noisy-channel approach to depth-charge illusions.
\newblock \emph{Cognition}, 232:105346.

\end{thebibliography}
\bibliographystyle{acl_natbib}
\newpage

\appendix
\section{Experiment materials}
\label{app:materials}
In Table \ref{tab:appendix_sentences} are all the sentences for each of our structures.

\begin{table*}[t!]
\centering
\scriptsize
\begin{tabular}{p{0.45\textwidth} p{0.45\textwidth}}
\midrule
\multicolumn{2}{c}{\textbf{Subject/Object (GP) (Amouyal et al., 2025)}} \\
\midrule
1. While the secretary typed the memo neared completion. & 2. While Janet baked the bread rose in the oven. \\
3. While the explorer paddled the canoe headed toward a waterfall. & 4. While the public cheered the team left the restaurant. \\
5. While the cowboy rode the horse sweated profusely. & 6. While the cleaner mopped the floor was filled with stains. \\
7. While Tom grilled the hot dog began to burn. & 8. While the chef cooked the meal impressed the couple. \\
9. While the architect drew the building represented modern times style. & 10. While the child finished the homework waited on the table. \\
11. While the chef stirred the soup boiled over. & 12. While the student knitted the sweater sold to the highest bidder. \\
13. While the tourist explored the tunnel echoed with mysterious sounds. & 14. While the astronomer observed the comet lit up the room. \\
15. While the woman drank the water spilled on the floor. & 16. While the players started the game bored the children. \\
17. While the snake swallowed the frog kicked vigorously. & 18. While the professor taught the students looked at the board. \\
19. While the lion attacked the baboon screamed in terror. & 20. While the pianist practiced the melody echoed through the hall. \\
21. While the maid dusted the picture tipped over. & 22. While the couple left the bar buzzed with activity. \\
23. While the teacher counted the children formed a line. & 24. While the gardener harvested the tomatoes hanged on the vine. \\
25. While the champion raced the challenger stumbled and fell. & 26. While the horse pulled the cart moved silently. \\
27. While Jerry played the violin went out of tune. & 28. While the man hunted the deer ran into the woods. \\
29. While the girl painted the rainbow slowly faded outside. & 30. While the skipper sailed the boat veered off course. \\
31. While Kendra parked the van bumped the curb. & 32. While the orchestra performed the symphony played on the radio. \\
33. While Angela cleaned the dog stood in the yard. & 34. While the bridesmaid ordered the dress got delivered. \\
35. While the sailor smoked the pipe glowed brightly. & 36. While Susan wrote the letter fell off the table. \\
37. While the tourist filmed the dancer blocked the sidewalk. & 38. While the farmer steered the tractor pulled the car. \\
39. While the athlete wrestled the opponent shouted insults. & 40. While the lawyer studied the contract lay on the roll-top desk. \\
41. While the warrior fought the enemy retreated. & 42. While the clown juggled the balls fell on the ground. \\
43. While Harry chewed the steak fell to the floor. & 44. While Anne vacuumed the rug lost its colors. \\
45. While Bill ate the turkey sat on the table. &  \\
\bottomrule
\end{tabular}
\begin{tabular}{p{0.3\textwidth} p{0.3\textwidth} p{0.3\textwidth}}
\multicolumn{3}{c}{\textbf{NP/VP (GP)}} \\
\midrule
1. The complex houses married soldiers. & 2. The devoted stage the protest. & 3. The poor taste the soup. \\
4. The strong arm the cannons. & 5. The mighty power the engines. & 6. The free press the grapes. \\
7. The official addresses the crowd. & 8. The brave mount the horses. & 9. The clean sweep the floors. \\
10. The strong bear the load. & 11. The poor harvest the crops. & 12. The private school the recruits. \\
13. The experienced coach the team. & 14. The young plant the trees. & 15. The public transport the supplies. \\
16. The professional cook the meals. & 17. The weak lift the boards. & 18. The blind date the new girl. \\
19. The graceful dance the waltz. & 20. The old record the race. & 21. The weak point the way. \\
22. The brave face the danger. & 23. The poor light the lanterns. & 24. The hard drive the cattle. \\
25. The orderly file the document. & 26. The rich soil the gardens. & 27. The quick fix the leaks. \\
28. The young fish salmon. & 29. The blind spot the birds. & 30. The strong drink the ale. \\
31. The skilled hand the control. & 32. The sick leave the building. & 33. The bold move the furniture. \\
34. The designated head the committee. & 35. The simple answer the questionnaire. & 36. The slow burn the letters. \\
37. The clever fool the guards. & 38. The just cause the revolt. &  
39. The seasoned grill the steaks.\\ & 40. The sharp edge the tiles. &  \\
\midrule
\end{tabular}
\begin{tabular}{p{0.45\textwidth} p{0.45\textwidth}}
\multicolumn{2}{c}{\textbf{Reduced relative (GP)}} \\
\midrule
1. The chef hired last month worked overtime. & 2. The doctor tested last week gave out prescriptions. \\ 3. The friends invited to the gala mingled with guests. & 4.
The committee selected this morning met the candidates. \\ 5. The editor published this month received praise. & 6. The leaders inspired last year implemented reforms. \\
7. The teachers praised in the review felt encouraged. & 8. The manager blamed after the incident left the office. \\ 9. The client thanked at the reception booked a boat trip. & 10.
The staff appointed last week started work. \\ 11. The soldiers attacked at midnight held their ground. & 12. The author criticized last week defended his work. \\
13. The scientist nominated for the award gave a speech. & 14. The workers replaced during the strike returned to their posts. \\ 15. The company financed this quarter expanded its reach. & 16. The coach trained overseas shouted at its team. \\ 17. The professor reviewed last year received a prize. & 18. The guest met at the conference called yesterday. \\
19. The officers promoted this year attended the ceremony. & 20. The firefighters instructed during orientation conducted drills. \\ 21. The lawyers sued last year took on new cases. &
22. The directors elected in 2020 resigned. \\ 23. The mentors guided through the program offered feedback. & 24. The doctors treated at the clinic improved quickly. \\
25. The manager interviewed for the position seemed qualified. & 26. The headhunter recruited last week looked for candidates. \\ 27. The florist sent the flowers was pleased. & 28.
The artists commissioned for the sculpture received awards. \\ 29. The department merged last year diversified operations. & 30. The screen displayed in the mall broke suddenly \\
31. The policemen dispatched yesterday found violations. & 32. The students surveyed for feedback passed the class. \\ 33. The fugitive wanted in Europe left the US. & 34.
The lawyer consulted for free gained experience. \\ 35. The guides accompanied on the tour provided assistance. & 36. The father caught on camera drove too fast. \\
37. The teachers rewarded with bonuses improved morale. & 38. The policeman arrested during the raid stole weapons. \\
39. The robots manufactured last quarter malfunctioned. & 40. The cat captured in the forest slept on the couch. \\
\bottomrule
\end{tabular}
\end{table*}

\begin{table*}[t!]
\centering
\scriptsize
\begin{tabular}{p{0.48\textwidth} p{0.48\textwidth}}
\midrule
\multicolumn{2}{c}{\textbf{NP/S (GP)}} \\
\midrule
1. The student forgot the solution was in the back of the book. & 2. The scientist proved the theory was incorrect after experiments. \\
3. The chef remembered the recipe required two hours of cooking. & 4. The president announced the policy was approved unanimously. \\
5. The coach discovered the player tried to show off all the time. & 6. The tennisman conceded the point was beautiful. \\
7. The student understood the question was too hard. & 8. The president declared war was not an option. \\
9. The musician heard the song was good. & 10. The filmmaker showed the movie was bad. \\
11. The reporter believed the story turned out problematic. & 12. The guard recognized the visitor was armed. \\
13. She recalled the conversation ended abruptly. & 14. The couple regretted the wedding lasted only 2 hours. \\
15. Tom noticed the murderer left town. & 16. The reporter repeated the claim was false. \\
17. The policeman saw the lights were off. & 18. The driver forgot the car slowed down at night. \\
19. The journalist shared the scoop was fake. & 20. The guard remembered the code changed last night. \\
21. The manager noticed the report was overdue. & 22. The tourist discovered the bridge was closed for the holiday.  \\
23. The art curator explained the painting sold for 4 million dollars. & 24. The undergrad understood the formula was too complicated for him. \\
25. The driver knew the road was slippery. & 26. The manager heard the plan fell through. \\
27. The professor realized the experiment failed due to contamination. & 28. The analyst believed the numbers were wrong. \\
29. The plaintiff accepted the verdict was unfair. & 30. The painter recalled the model stood him up. \\
31 The committee found the proposal was flawed. & 32. The guard saw the diamond disappeared overnight. \\
33. The editor mentioned the errors were corrected.  & 34. The critic knew the song was very popular. \\
35. The archaeologist revealed the artifact broke during extraction. & 36. The student learned the lesson was canceled. \\
37. The new hire learned Spanish was mandatory for client meetings. & 38. The experiment proved the claim was incorrect. \\
39. The detective disclosed the evidence was only recently found.  & 40. The musician recognized the song was catchy. \\
\midrule
\multicolumn{2}{c}{\textbf{Double center embedding}} \\
\midrule
1. The man that the teacher that the student liked called sat. & 2. The sculptor that the model that the photographer interviewed inspired rested. \\
3. The doctor that the nurse that the patient trusted praised arrived. & 4. The programmer that the tester that the manager supervised debugged coughed. \\
5. The neighbor that the parent that the child thanked invited waved. & 6. The knight that the squire that the bard praised served bowed. \\
7. The author that the editor that the reviewer praised interviewed smiled. & 8. The violinist that the conductor that the critic praised directed bowed. \\
9. The scientist that the assistant that the intern admired warned laughed. & 10. The senator that the lobbyist that the voter called persuaded spoke. \\
11. The actor that the director that the critic praised hired performed. & 12. The mountaineer that the guide that the photographer hired led slipped. \\
13. The chef that the waiter that the customer thanked helped cooked. & 14. The banker that the attorney that the journalist interviewed represented coughed. \\
15. The mayor that the journalist that the photographer called questioned spoke. & 16. The firefighter that the medic that the witness thanked rescued smiled. \\
17. The engineer that the manager that the investor trusted promoted resigned. & 18. The merchant that the broker that the customer trusted advised laughed. \\
19. The singer that the producer that the fan liked hugged performed. & 20. The king that the advisor that the noble criticized guided wept. \\
21. The researcher that the lab assistant that the supervisor instructed assisted published. & 22. The magician that the assistant that the child admired helped bowed. \\
23. The soldier that the officer that the reporter questioned commanded marched. & 24. The umpire that the commentator that the player applauded criticized frowned. \\
25. The athlete that the coach that the scout evaluated trained competed. & 26. The biologist that the curator that the volunteer assisted guided sneezed. \\
27. The farmer that the merchant that the customer paid supplied harvested. & 28. The pianist that the tutor that the student hired trained smiled. \\
29. The painter that the curator that the visitor admired guided smiled. & 30. The carpenter that the foreman that the architect hired managed hammered. \\
31. The novelist that the agent that the critic praised represented spoke. & 32. The minister that the advisor that the secretary contacted briefed coughed. \\
33. The dancer that the choreographer that the spectator cheered instructed spun. & 34. The hiker that the ranger that the tourist consulted escorted slipped. \\
35. The lawyer that the judge that the journalist questioned summoned coughed. & 36. The clerk that the supervisor that the auditor evaluated trained yawned. \\
37. The pilot that the mechanic that the supervisor thanked assisted rested. & 38. The courier that the dispatcher that the customer called disliked sighed. \\
39. The captain that the sailor that the tourist praised assisted yawned. & 40. The monk that the scholar that the pilgrim visited guided meditated. \\
\midrule
\multicolumn{2}{c}{\textbf{Depth charge}} \\
\midrule
1. No head injury is too trivial to be ignored. & 2. No goal is too distant to be put aside. \\
3. No detail is too minor to be overlooked. & 4. No emotion is too fleeting to be set aside. \\
5. No question is too simple to be dismissed. & 6. No truth is too uncomfortable to be left out. \\
7. No error is too small to be disregarded. & 8. No victory is too small to be minimized. \\
9. No concern is too insignificant to be quashed. & 10. No crime is too petty to be pardoned. \\
11. No risk is too small to be unadressed. & 12. No promise is too minor to be broken. \\
13. No task is too easy to be neglected. & 14. No wound is too shallow to be left untreated. \\
15. No symptom is too mild to be overlooked. & 16. No project is too useless to be neglected. \\
17. No complaint is too trivial to be put aside. & 18. No adventure is too dangerous to be ruled out. \\
19. No warning is too faint to be discounted. & 20. No witness is too unreliable to be discounted. \\
21. No opportunity is too slight to be skipped. & 22. No award is too minor to be looked past. \\
23. No act of kindness is too small to be minimized. & 24. No lesson is too unimportant to be skipped. \\
25. No effort is too minor to be omitted. & 26. No skill is too elementary to be snubbed. \\
27. No idea is too unconventional to be abandoned. & 28. No cause is too minor to be ditched. \\
29. No contribution is too insignificant to be snubbed. & 30. No experiment is too basic to be left out. \\
31. No suggestion is too outlandish to be forgotten. & 32. No child is too mean to be ostracized. \\
33. No gesture is too subtle to be waved off. & 34. No lecture is too boring to be unattended. \\
35. No tradition is too new to be abandoned. & 36. No painting is too ugly to be destroyed. \\
37. No achievement is too modest to be unrecognized. & 38. No culture is too bizarre to be rejected. \\
39. No dream is too ambitious to be dropped. & 40. No job is too small to be dismissed. \\
\bottomrule
\end{tabular}
\end{table*}

\begin{table*}[t!]
\centering
\scriptsize
\begin{tabular}{p{0.45\textwidth} p{0.45\textwidth}}
\toprule
\multicolumn{2}{c}{\textbf{Similarity interference (Gordon et al., 2001)}} \\
\midrule
1. The banker that the barber praised climbed the mountain just outside of town before it snowed. & 2. The actor that the director thanked worked in many hit movies before 1990. \\ 
3. The admiral that the general advised reminisced nostalgically before the trip got underway. &
4. The dancer that the reporter phoned cooked the pork chops in their own juices on New Year's Eve.\\ 
5. The poet that the painter inspired wrote an autobiography after their friendship became well known. & 6. The coach that the referee criticized talked publicly about the incident after the game. \\
7. The architect that the fireman liked dominated the conversation while the game was on television. & 8. The chef that the cashier distrusted called for help after the restaurant closed.\\ 9. The lawyer that the client interviewed had a very small office. &
10. The waiter that the broker despised drove the sports car home from work that evening. \\ 11. The aunt that the child amused made paper dolls out of the newspaper. & 12. The plumber that the electrician called drove a grey truck. \\
13. The detective that the secretary disliked clipped the coupons out with the dull scissors. & 14. The violinist that the conductor complimented performed at Carnegie Hall for two weeks. \\ 15. The salesman that the accountant contacted spoke very quickly. & 16. The judge that the doctor ignored watched the special about Colombian drug dealers on the nightly news. \\ 17. The teacher that the student questioned wrote a long science fiction novel during the summer vacation. & 18. The clown that the magician entertained was a star. \\
19. The robber that the mailman insulted read the newspaper article about the fire. & 20. The editor that the author recommended changed jobs after a new merger was announced.\\ 21. The clerk that the traveler helped worked in a large foreign bank. & 22.
The governor that the comedian admired answered the telephone in the fancy restaurant. \\ 23. The tailor that the customer described worked in a small building near the bus station. & 24. The gardener that the homeowner envied was very friendly. \\
\midrule
\end{tabular}
\caption{Experimental sentences from the hardest conditions for each syntactic structure.}
\label{tab:appendix_sentences}
\end{table*} 
\section{Prompts used}
\label{app:prompt_ise}

Figure \ref{fig:prompt_1} shows an example prompt with each system prompt. The other prompts differ in the order the examples within the prompt. All our prompts can be find in our codebase at \emph{Anonymised}.

\begin{figure*}
\colorbox{gray!10}{
\begin{minipage}{15cm}
\texttt{You are a linguistic experiment subject. You will be presented with a sentence, and will need to answer a reading comprehension question. You will need to select an option amongst the proposed answers.
 \\Here are a few examples of questions and relevant answers:
 \\
 \\The doctor that she called checked on the patient yesterday.
 \\
 \\Answer with Yes or No:
 \\Did she call the doctor?
 \\My answer is: Yes
 \\
 \\
 \\The sailor that John punished stayed in his room.
 \\
 \\Answer with Yes or No:
 \\Did John stay in his room?
 \\My answer is: No
 \\
 \\
 \\The professor that emailed the surgeon was stuck on a case.
 \\
 \\Answer with Yes or No:
 \\Did the surgeon email the professor?
 \\My answer is: No
 \\
 \\
 \\The driver that saved the cyclist went back home.
 \\
 \\Answer with Yes or No:
 \\Did the driver go back home?
 \\My answer is: Yes
 }
\end{minipage}}
\caption{Example of the first system prompt}
\label{fig:prompt_1}
\end{figure*}

\begin{figure*}
\colorbox{gray!10}{
\begin{minipage}{15cm}
    \texttt{You will answer a reading comprehension question about a sentence.
\\Here are a few examples of questions and correct answers:
\\
\\The singer that hired the guitarist arrived to the concert early.
\\
\\Answer with Yes or No:
\\Did the singer hire the guitarist?
\\My answer is: Yes
\\
\\
\\The doctor that the nurse called checked on the patient yesterday.
\\
\\Answer with Yes or No:
\\Did the nurse call the doctor?
\\My answer is: Yes
\\
\\
\\The professor that emailed Matt was stuck on a case.
\\
\\Answer with Yes or No:
\\Did Matt email the professor?
\\My answer is: No
\\
\\
\\The teacher that helped you graded the papers on the weekend.
\\
\\Answer with Yes or No:
\\Did you grade the papers?
\\My answer is: No
}
\end{minipage}}
\caption{Example of the second system prompts}
\label{fig:prompt_2}
\end{figure*}
\section{Full results}
\label{app:full_res}

Table \ref{tab:appendix_all_models_full_res} shows the results on the target and baseline conditions for all the models.

\begin{table*}[t!]
\centering
\scriptsize
\begin{tabular}{lcccccccccccccc}
\toprule
Model & \multicolumn{2}{c}{Subj/Obj. GP} & \multicolumn{2}{c}{NP/S GP} & \multicolumn{2}{c}{NP/VP GP} & \multicolumn{2}{c}{Red. rel.} & \multicolumn{2}{c}{Depth charge} & \multicolumn{2}{c}{Double center} & \multicolumn{2}{c}{Interference} \\
 & Targ. & Base & Targ. & Base & Targ. & Base & Targ. & Base & Targ. & Base & Targ. & Base & Targ. & Base \\
\midrule
Qwen.0.6B & 0.36 & 0.36 & 0.39 & 0.35 & 0.65 & 0.64 & 0.40 & 0.46 & 0.51 & 0.51 & 0.52 & 0.77 & 0.52 & 0.51 \\
Qwen.1.7B & 0.09 & 0.21 & 0.21 & 0.05 & 0.41 & 0.46 & 0.32 & 0.60 & 0.64 & 0.55 & 0.49 & 0.89 & 0.62 & 0.52 \\
Qwen.4B & 0.03 & 0.15 & 0.17 & 0.21 & 0.27 & 0.42 & 0.18 & 0.58 & 0.55 & 0.83 & 0.55 & 0.89 & 0.68 & 0.70 \\
Qwen.8B & 0.28 & 0.22 & 0.33 & 0.31 & 0.42 & 0.43 & 0.38 & 0.74 & 0.56 & 0.87 & 0.66 & 0.90 & 0.77 & 0.87 \\
Qwen.14B & 0.12 & 0.12 & 0.10 & 0.18 & 0.22 & 0.42 & 0.34 & 0.79 & 0.79 & 0.86 & 0.79 & 0.88 & 0.91 & 0.87 \\
Qwen.32B & 0.05 & 0.09 & 0.16 & 0.34 & 0.28 & 0.62 & 0.53 & 0.86 & 0.52 & 0.76 & 0.69 & 0.84 & 0.87 & 0.80 \\ \hline
Deepseek-1.5B & 0.38 & 0.42 & 0.39 & 0.34 & 0.39 & 0.32 & 0.40 & 0.44 & 0.69 & 0.72 & 0.59 & 0.84 & 0.43 & 0.41 \\
Deepseek-7B & 0.32 & 0.40 & 0.44 & 0.30 & 0.41 & 0.43 & 0.53 & 0.69 & 0.78 & 0.68 & 0.25 & 0.85 & 0.68 & 0.74 \\ 
Deepseek-14B & 0.15 & 0.12 & 0.07 & 0.09 & 0.25 & 0.28 & 0.25 & 0.53 & 0.74 & 0.84 & 0.66 & 0.87 & 0.65 & 0.71 \\ \hline
Gemma-1B & 0.40 & 0.39 & 0.38 & 0.36 & 0.43 & 0.43 & 0.39 & 0.40 & 0.65 & 0.58 & 0.44 & 0.57 & 0.41 & 0.44 \\
Gemma-1B-Ins. & 0.40 & 0.49 & 0.34 & 0.24 & 0.52 & 0.49 & 0.37 & 0.44 & 0.64 & 0.46 & 0.20 & 0.30 & 0.28 & 0.39 \\
Gemma-4B & 0.35 & 0.34 & 0.37 & 0.34 & 0.38 & 0.42 & 0.34 & 0.38 & 0.55 & 0.54 & 0.35 & 0.48 & 0.35 & 0.42 \\
Gemma-4B-Ins. & 0.40 & 0.50 & 0.36 & 0.34 & 0.37 & 0.41 & 0.34 & 0.38 & 0.54 & 0.54 & 0.04 & 0.13 & 0.35 & 0.40 \\
Gemma-12B & 0.23 & 0.20 & 0.23 & 0.26 & 0.26 & 0.33 & 0.25 & 0.36 & 0.62 & 0.59 & 0.44 & 0.58 & 0.42 & 0.41 \\
Gemma-12B-Ins. & 0.12 & 0.13 & 0.18 & 0.15 & 0.20 & 0.30 & 0.24 & 0.75 & 0.64 & 0.78 & 0.60 & 0.61 & 0.70 & 0.78 \\
Gemma-27B & 0.14 & 0.13 & 0.15 & 0.16 & 0.16 & 0.24 & 0.18 & 0.31 & 0.38 & 0.44 & 0.38 & 0.50 & 0.27 & 0.28 \\
Gemma-27B-Ins. & 0.07 & 0.25 & 0.20 & 0.24 & 0.32 & 0.60 & 0.31 & 0.95 & 0.53 & 0.80 & 0.14 & 0.22 & 0.82 & 0.87 \\ \hline
Llama-1B & 0.33 & 0.34 & 0.35 & 0.33 & 0.30 & 0.29 & 0.32 & 0.34 & 0.80 & 0.74 & 0.33 & 0.83 & 0.35 & 0.32 \\
Llama-1B-Ins. & 0.40 & 0.41 & 0.47 & 0.40 & 0.24 & 0.20 & 0.12 & 0.18 & 0.75 & 0.70 & 0.26 & 0.78 & 0.28 & 0.31 \\
Llama-3B & 0.30 & 0.37 & 0.30 & 0.24 & 0.38 & 0.37 & 0.37 & 0.43 & 0.53 & 0.61 & 0.59 & 0.80 & 0.34 & 0.44 \\
Llama-3B-Ins. & 0.42 & 0.52 & 0.49 & 0.39 & 0.61 & 0.55 & 0.63 & 0.70 & 0.38 & 0.40 & 0.41 & 0.80 & 0.51 & 0.80 \\
Llama-11B & 0.47 & 0.43 & 0.56 & 0.52 & 0.66 & 0.77 & 0.72 & 0.86 & 0.38 & 0.39 & 0.62 & 0.75 & 0.82 & 0.90 \\
Llama-11B-Ins. & 0.50 & 0.41 & 0.62 & 0.54 & 0.72 & 0.85 & 0.79 & 0.93 & 0.33 & 0.34 & 0.61 & 0.79 & 0.69 & 0.80 \\
Llama-90B & 0.25 & 0.20 & 0.29 & 0.42 & 0.40 & 0.57 & 0.39 & 0.87 & 0.43 & 0.62 & 0.88 & 0.80 & 0.93 & 0.91 \\
Llama-90B-Ins. & 0.17 & 0.15 & 0.24 & 0.39 & 0.37 & 0.59 & 0.36 & 0.93 & 0.43 & 0.66 & 0.90 & 0.82 & 0.98 & 0.96 \\ \hline
GPT-4o & 0.42 & 0.42 & 0.47 & 0.58 & 0.59 & 0.78 & 0.88 & 0.98 & 0.32 & 0.78 & 0.90 & 0.88 & 0.98 & 0.99 \\
GPT-4.1 & 0.40 & 0.27 & 0.57 & 0.71 & 0.63 & 0.88 & 0.76 & 0.98 & 0.63 & 0.99 & 0.90 & 0.88 & 1.00 & 1.00 \\
o3 & 0.49 & 0.37 & 0.56 & 0.76 & 0.66 & 0.89 & 0.95 & 1.00 & 0.64 & 1.00 & 0.98 & 0.95 & 1.00 & 1.00 \\
o3-mini & 0.18 & 0.63 & 0.16 & 0.52 & 0.38 & 0.85 & 0.71 & 1.00 & 0.49 & 0.91 & 0.97 & 0.95 & 1.00 & 1.00 \\
o4-mini & 0.28 & 0.39 & 0.56 & 0.66 & 0.52 & 0.87 & 0.82 & 0.99 & 0.57 & 0.96 & 0.69 & 0.88 & 0.99 & 0.98 \\
GPT-5 & 0.32 & 0.27 & 0.45 & 0.53 & 0.45 & 0.80 & 0.65 & 0.98 & 0.83 & 0.94 & 0.98 & 0.95 & 1.00 & 1.00 \\
\bottomrule
\end{tabular}
\caption{Accuracy of all models on all structures. For each structure, Target (Targ.) and Baseline (Base) conditions are shown side by side.}
\label{tab:appendix_all_models_full_res}
\end{table*}
\section{Difference to humans}
\label{app:diff_humans}

In Table \ref{tab:human_differences} we show the absolute difference in accuracy between humans and LLMs on various structures.

\begin{table}
\centering
\small
\begin{tabular}{lcccc}
\toprule
Model & Avg GP & Depth ch. & Dbl. ctr. & Interf. \\
\midrule
Qwen-0.6B & 0.200 & 0.230 & 0.198 & 0.151 \\
Qwen-1.7B & 0.113 & 0.360 & 0.167 & 0.251 \\
Qwen-4B & 0.138 & 0.270 & 0.228 & 0.311 \\
Qwen-8B & 0.113 & 0.280 & 0.338 & 0.401 \\
Qwen-14B & 0.081 & 0.510 & 0.468 & 0.541 \\
Qwen-32B & 0.107 & 0.240 & 0.367 & 0.501 \\ \hline
Deepseek-1.5B & 0.140 & 0.410 & 0.267 & 0.061 \\
Deepseek-7B & 0.167 & 0.500 & 0.073 & 0.311 \\
Deepseek-14B & 0.119 & 0.460 & 0.338 & 0.281 \\ \hline
Gemma-1B & 0.155 & 0.370 & 0.117 & 0.041 \\
Gemma-1B-Ins. & 0.173 & 0.360 & 0.122 & 0.089 \\
Gemma-4B & 0.140 & 0.270 & 0.027 & 0.019 \\
Gemma-4B-Ins. & 0.148 & 0.260 & 0.283 & 0.019 \\
Gemma-12B & 0.102 & 0.340 & 0.117 & 0.051 \\
Gemma-12B-Ins. & 0.081 & 0.360 & 0.277 & 0.331 \\
Gemma-27B & 0.104 & 0.100 & 0.057 & 0.099 \\
Gemma-27B-Ins. & 0.101 & 0.250 & 0.182 & 0.451 \\ \hline
Llama1B & 0.115 & 0.520 & 0.008 & 0.019 \\
Llama1B-Instruct & 0.198 & 0.470 & 0.062 & 0.089 \\
Llama3B & 0.103 & 0.250 & 0.267 & 0.029 \\
Llama3B-Instruct & 0.279 & 0.100 & 0.087 & 0.141 \\
Llama11B & 0.344 & 0.100 & 0.297 & 0.451 \\
Llama11B-Ins. & 0.399 & 0.050 & 0.287 & 0.321 \\
Llama90B & 0.092 & 0.150 & 0.557 & 0.561 \\
Llama90B-Ins. & 0.084 & 0.150 & 0.578 & 0.611 \\
GPT-4o & 0.332 & 0.040 & 0.578 & 0.611 \\
GPT-4.1 & 0.332 & 0.350 & 0.578 & 0.631 \\
o3 & 0.407 & 0.360 & 0.657 & 0.631 \\
o3-mini & 0.168 & 0.210 & 0.647 & 0.631 \\
o4-mini & 0.287 & 0.290 & 0.367 & 0.621 \\
GPT-5 & 0.209 & 0.55 & 0.658 & 0.631 \\
\bottomrule
\end{tabular}
\caption{Absolute accuracy differences from human performance for each model.}
\label{tab:human_differences}
\end{table}
\section{Full results directionality}
\label{app:full_direc}

In Table \ref{tab:directionality_pattern} we can see the directionality for all the models on all the structures.

\begin{table*}
\centering
\small
\begin{tabular}{l|cccc|ccc}
\toprule
Model & Subj/Obj GP & NP/S GP & NP/VP GP & Red. rel. GP & Depth charge & Double center & Interference \\
\midrule
Qwen-0.6B & \xmark & \xmark & \xmark & \cmark & \xmark & \cmark & \xmark \\
Qwen-1.7B & \cmark & \cmark & \xmark & \cmark & \xmark & \cmark & \xmark \\
Qwen-4B & \cmark & \cmark & \cmark & \cmark & \cmark & \cmark & \cmark \\
Qwen-8B & \xmark & \cmark & \xmark & \cmark & \cmark & \cmark & \cmark \\
Qwen-14B & \xmark & \cmark & \cmark & \cmark & \cmark & \cmark & \xmark \\
Qwen-32B & \cmark & \cmark & \cmark & \cmark & \cmark & \cmark & \xmark \\ \hline
Deepseek-1.5B & \cmark & \xmark & \xmark & \cmark & \cmark & \cmark & \xmark \\
Deepseek-7B & \cmark & \cmark & \xmark & \cmark & \xmark & \cmark & \cmark \\
Deepseek-14B & \xmark & \cmark & \cmark & \cmark & \cmark & \cmark & \cmark \\ \hline
Gemma-1B & \xmark & \xmark & \xmark & \cmark & \xmark & \cmark & \cmark \\
Gemma-1B-Ins. & \cmark & \xmark & \xmark & \cmark & \xmark & \cmark & \cmark \\
Gemma-4B & \xmark & \cmark & \xmark & \cmark & \xmark & \cmark & \cmark \\
Gemma-4B-Ins. & \cmark & \cmark & \xmark & \cmark & \xmark & \cmark & \cmark \\
Gemma-12B & \xmark & \cmark & \cmark & \cmark & \xmark & \cmark & \xmark \\
Gemma-12B-Ins. & \cmark & \cmark & \xmark & \cmark & \cmark & \cmark & \cmark \\
Gemma-27B & \xmark & \cmark & \cmark & \cmark & \cmark & \cmark & \cmark \\
Gemma-27B-Ins. & \cmark & \cmark & \cmark & \cmark & \cmark & \cmark & \cmark \\ \hline
Llama-1B & \cmark & \xmark & \xmark & \cmark & \xmark & \cmark & \xmark \\
Llama-1B-Ins. & \cmark & \xmark & \xmark & \cmark & \xmark & \cmark & \cmark \\
Llama-3B & \cmark & \xmark & \xmark & \cmark & \cmark & \cmark & \cmark \\
Llama-3B-Ins. & \cmark & \xmark & \xmark & \cmark & \cmark & \cmark & \cmark \\
Llama-11B & \xmark & \cmark & \xmark & \cmark & \cmark & \cmark & \cmark \\
Llama-11B-Ins. & \xmark & \cmark & \xmark & \cmark & \cmark & \cmark & \cmark \\
Llama-90B & \xmark & \cmark & \cmark & \cmark & \cmark & \xmark & \xmark \\
Llama-90B-Ins. & \xmark & \cmark & \cmark & \cmark & \cmark & \xmark & \xmark \\ \hline
GPT-4o & \xmark & \cmark & \cmark & \cmark & \cmark & \xmark & \cmark \\
GPT-4.1 & \xmark & \cmark & \cmark & \cmark & \cmark & \xmark & \xmark \\
o3 & \xmark & \cmark & \cmark & \cmark & \cmark & \xmark & \xmark \\
o3-mini & \cmark & \cmark & \cmark & \cmark & \cmark & \xmark & \xmark \\
o4-mini & \cmark & \cmark & \cmark & \cmark & \cmark & \cmark & \xmark \\
GPT-5 & \xmark & \cmark & \cmark & \cmark & \cmark & \xmark & \xmark  \\
\bottomrule
\end{tabular}
\caption{Directionality pattern for each model on all structures. Crosses mark cases for which the accuracy on the baseline sentences is lower than on the difficult, and check signs marks the opposite.}
\label{tab:directionality_pattern}
\end{table*}

\end{document}